%% file: main.tex
\documentclass[10pt,twocolumn,letterpaper]{article}

\usepackage{cvpr}              

\input{preamble}
\definecolor{cvprblue}{rgb}{0.21,0.49,0.74}
\usepackage[pagebackref,breaklinks,colorlinks,allcolors=cvprblue, bookmarks=false]{hyperref}

\usepackage{booktabs}
\usepackage{graphicx}
\usepackage{caption,subcaption}
\usepackage{tikz}
\usepackage{algorithm}
\usepackage{algpseudocode}
\usepackage{appendix}

\def\paperID{16254} 
\def\confName{CVPR}
\def\confYear{2025}

\title{MEt3R: Measuring Multi-View Consistency in Generated Images}

\author{Mohammad Asim\textsuperscript{1} \quad Christopher Wewer\textsuperscript{1} \quad Thomas Wimmer\textsuperscript{1,2} \quad Bernt Schiele\textsuperscript{1} \quad Jan Eric Lenssen\textsuperscript{1}\\
{\normalsize \textsuperscript{1}Max Planck Institute for Informatics, Saarland Informatics Campus \quad
\textsuperscript{2}ETH Zurich}\\
{\tt\small \{masim, jlenssen\}@mpi-inf.mpg.de}
}

\begin{document}

\twocolumn[{%
\renewcommand\twocolumn[1][]{#1}%

\maketitle
\begin{center}
\centering
\captionsetup{type=figure}
{
\centering
    \hspace*{\fill}%
    \raisebox{-0.5\height}{\includegraphics[width=.70\textwidth]{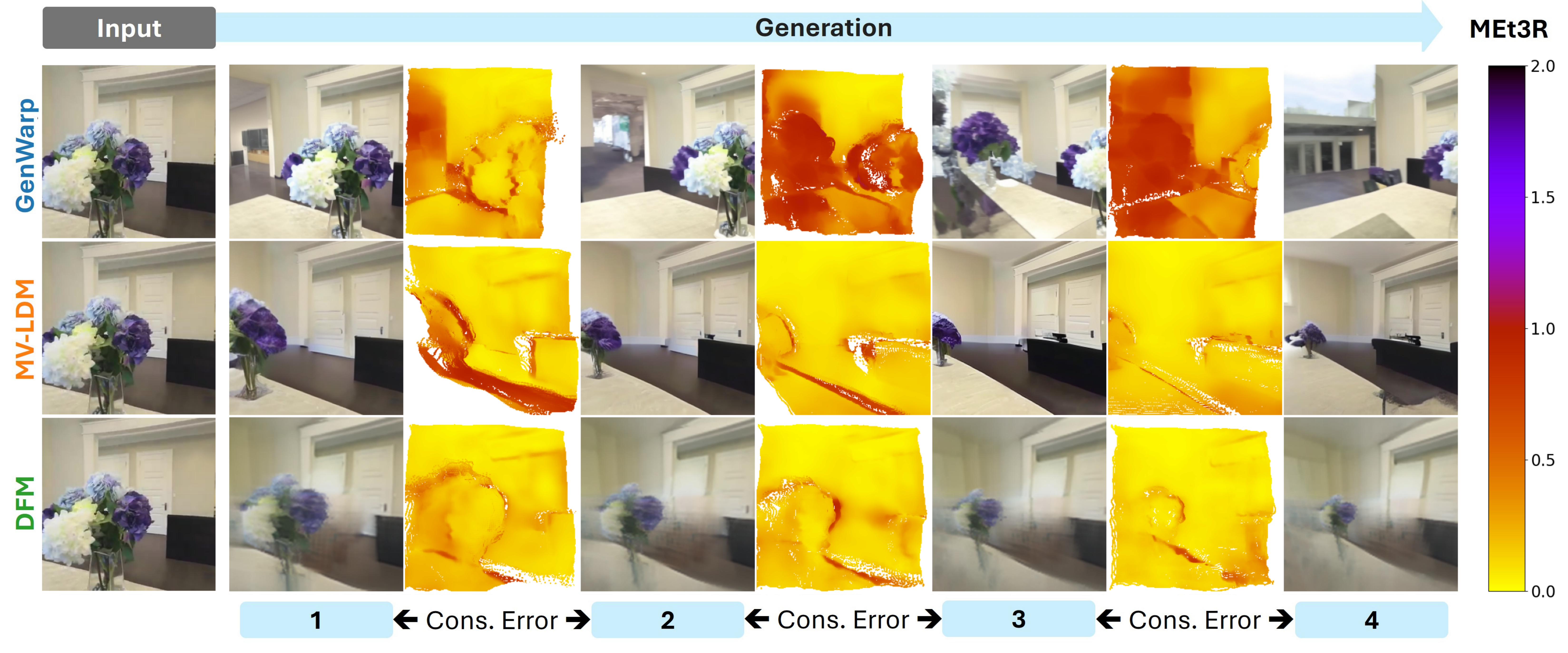}}
    \hspace*{\fill}%
    {\raisebox{-0.5\height}{\includegraphics[width=.29\textwidth]{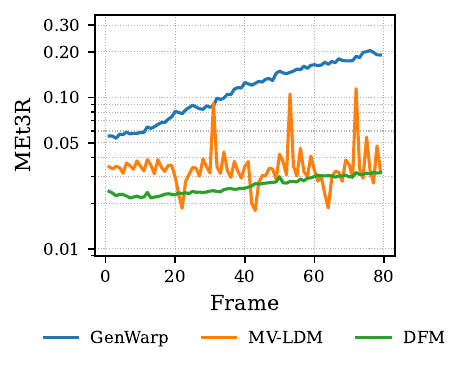}}}
    \hspace*{\fill}%
}
\captionof{figure}{We introduce \textbf{MEt3R}, a metric for multi-view consistency between pairs of generated images, which is independent of image quality and content and does not require camera poses. \textbf{Left}: generated images from different generative models, conditioned on the first frame, with MEt3R score map (Cons. Error) indicating levels of inconsistencies between consecutive images $i$ and $i+1$. \textbf{Right}: pair-wise consistency scores, evaluated for consecutive frames in a sliding window, averaged over multiple sequences. The pattern in MV-LDM's consistency clearly shows artifacts from using anchor frames that are generated first, highlighting the high signal-to-noise ratio of MEt3R. }
\label{fig:teaser}
\end{center}%
}]

\input{sec/0_abstract}    
\input{sec/1_intro}
\input{sec/2_relatedwork}

\input{sec/3_method}

\input{sec/4_experiments}

{
    \small
    \bibliographystyle{ieeenat_fullname}
    \bibliography{main}
}

\input{sec/X_suppl}

\end{document}

%% file: preamble.tex
\usepackage[table, dvipsnames]{xcolor}

%% file: sec/0_abstract.tex
\begin{abstract}
We introduce MEt3R, a metric for multi-view consistency in generated images. Large-scale generative models for multi-view image generation are rapidly advancing the field of 3D inference from sparse observations. However, due to the nature of generative modeling, traditional reconstruction metrics are not suitable to measure the quality of generated outputs and metrics that are independent of the sampling procedure are desperately needed. In this work, we specifically address the aspect of consistency between generated multi-view images, which can be evaluated independently of the specific scene. 
Our approach uses DUSt3R to obtain dense 3D reconstructions from image pairs in a feed-forward manner, which are used to warp image contents from one view into the other. Then, feature maps of these images are compared to obtain a similarity score that is invariant to view-dependent effects. Using MEt3R, we evaluate the consistency of a large set of previous methods for novel view and video generation, including our open, multi-view latent diffusion model. Code is available online: \url{geometric-rl.mpi-inf.mpg.de/met3r/}.
\end{abstract}

%% file: sec/1_intro.tex
\section{Introduction}
\label{sec:intro}
Generative models, such as diffusion~\citep{ddpm, ddim} or flow-based~\citep{liu2022flow} models, are trained to sample from a given data distribution, which makes them ideal candidates for stochastic inverse problems, such as reconstruction from incomplete information~\citep{reconfusion,gao2024cat3d,tewari2023forwarddiffusion}. However, they raise the inherent challenge that, for individual samples, no ground truth is available to measure the quality of generations with pairwise distance metrics. As a result, metrics such as FID~\citep{fid}, KID~\citep{binkowski2018demystifying}, and CMMD~\citep{jayasumana2024rethinking} have been proposed to measure the quality of generated images without the need for a paired ground truth.

\begin{figure*}[t]
    \centering
    \includegraphics[width=1\linewidth]{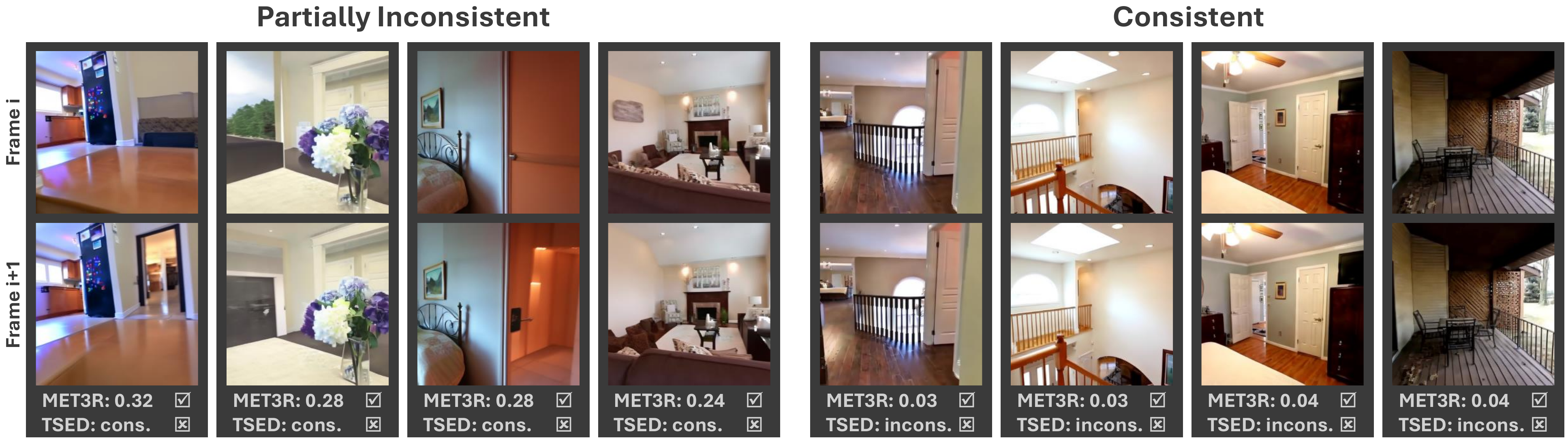}
    \caption{\textbf{Existing metrics.} A comparison between MEt3R and TSED~\citep{Yu2023PhotoconsistentNVS} scores obtained from individual image pairs generated by GenWarp~\cite{seo2024genwarp}. TSED misses obvious, partial multi-view inconsistencies and is biased to small violations of epipolar geometry. In contrast, MEt3R correctly captures clear 3D inconsistencies and is robust to insignificant artifacts almost invisible to the human eye.} 
    \label{fig:existing}
    \vspace{-0.3cm}
\end{figure*}

Recently, a trend is to repurpose video~\cite{ho2022video, blattmann2023align} and image~\cite{song2020score, rombach2022high} diffusion models for generation of 3D scenes and objects, by generating multiple views from different camera poses~\cite{gao2024cat3d, sparsefusion, voleti2025sv3d}, with or without given images as conditioning. Compared to direct generation of 3D representations~\cite{diffrf, ssdnerf, npcd}, such multi-view generative models can be trained on images and videos, and their pixel-aligned representation allows for more efficient models and better scalability. However, they have only a weak to non-existent inductive bias to produce actually 3D consistent results, which is of large importance for the subsequent lift into 3D. A reliable metric to evaluate the multi-view consistency of such generations is critically needed to advance these models further. Fortunately, similar to general image quality, 3D consistency between views can be evaluated without the existence of paired ground-truth data. Existing metrics such as TSED~\cite{Yu2023PhotoconsistentNVS}, though, fail to reliably perform such evaluation, as shown in Fig.~\ref{fig:existing}. In this work, we propose a metric to measure 3D consistency, which is independent of the specific scene and model used to generate the images, works under changing lighting conditions, does not require camera poses, is differentiable, and is a gradual measure of consistency instead of a binary one.

MEt3R utilizes DUSt3R~\cite{wang2023DUSt3R} to obtain dense reconstructions from image pairs in a common 3D space. It then projects features of one image into the view of the other using the reconstructed point maps and computes feature similarity between the obtained images. As feature extractors, DINO~\cite{caron2021emerging} + FeatUp~\cite{fu2024featup} are used to obtain high-resolution features from input images that are robust view-dependent effects, such as lighting, while preserving semantics and image-level structures, to quantify 3D consistency. We further introduce an open-source multi-view latent diffusion model (MV-LDM) to be used in our studies, which is able to generate good quality and consistent scenes. MEt3R is evaluated in different scenarios to validate its usefulness and robustness. It is used to benchmark existing methods that generate videos and multiple views of objects and scenes, with and without an intermediate 3D representation, as well as our MV-LDM. We show that MV-LDM performs well in the quality vs. consistency trade-off and find that MEt3R is a reliable metric that aligns well with the theoretical expectations of consistency among the different classes of scene generation methods. In contrast to previous metrics, it can distinguish perfectly consistent from almost consistent sequences and can robustly capture fine-grained changes in consistency over time. 

\noindent
In summary, our \textbf{contributions} include:
\begin{itemize}
    \item a simple yet effective metric for measuring 3D consistency of generated views without given camera poses,
    \item a comprehensive analysis of existing methods that generate videos and multiple views of objects and scenes, and
    \item an open-source multi-view latent diffusion model, which performs best in the quality vs. consistency trade-off.
\end{itemize}
Our code and models are publicly available.

%% file: sec/2_relatedwork.tex
\section{Related Work}
\label{sec:related_work}

We introduce a metric to evaluate the 3D consistency of multi-view generations. Thus, we review existing methods that generate multi-view representations of scenes and give an overview of existing quality metrics in this setting.
\vspace{-0.1cm}

\paragraph{Multi-view Generative Models.} \label{sec:mv-generative-models}
Recent success in 2D image generation using generative models like diffusion~\citep{rombach2022high} has sparked interest in generating 3D scenes. As the scarcity of high-quality training data and the complexity of 3D representations present a challenge for direct text-to-3D generative methods, recent methods explore repurposing image or video generation models as supervision signal or initialization for 3D generation~\citep{gao2024cat3d, seo2024genwarp, zero1to3, syncdreamer, shimvdream, genvs, reconfusion, sparsefusion, han2025vfusion3d, voleti2025sv3d, muller2024multidiff, tewari2023forwarddiffusion, Yu2023PhotoconsistentNVS, rombach2021geometryfree}.

3D-aware image generation methods can be grouped into methods for pose-conditioned single-view generation~\citep{seo2024genwarp, zero1to3, sparsefusion, Yu2023PhotoconsistentNVS, watson2022novel}, simultaneous multi-view image generation~\citep{gao2024cat3d, shimvdream, rombach2021geometryfree} and methods that use an internal 3D representation of the scene as prior for generation~\citep{syncdreamer, genvs, reconfusion, tewari2023forwarddiffusion, latentsplat}.
Further distinction can be made between models that are trained on single-asset 3D datasets~\citep{zero1to3, shimvdream, syncdreamer, watson2022novel}, such as Objaverse~\citep{objaverse}, and models trained on full 3D scenes~\citep{seo2024genwarp, rombach2021geometryfree, sparsefusion, Yu2023PhotoconsistentNVS, gao2024cat3d, genvs, reconfusion, tewari2023forwarddiffusion}.
Our introduced metric is agnostic to how images are generated.
In our experiments, we perform a comprehensive evaluation of consistency for images generated by openly available models, including those that model the joint distribution of input and single output views~\citep{seo2024genwarp, Yu2023PhotoconsistentNVS}, multiple output views~\citep{rombach2021geometryfree}, and methods that use an internal 3D representation~\citep{tewari2023forwarddiffusion} to enforce consistency. 

\label{sec:method}
\begin{figure*}[t]
    \centering
    \includegraphics[width=1\linewidth]{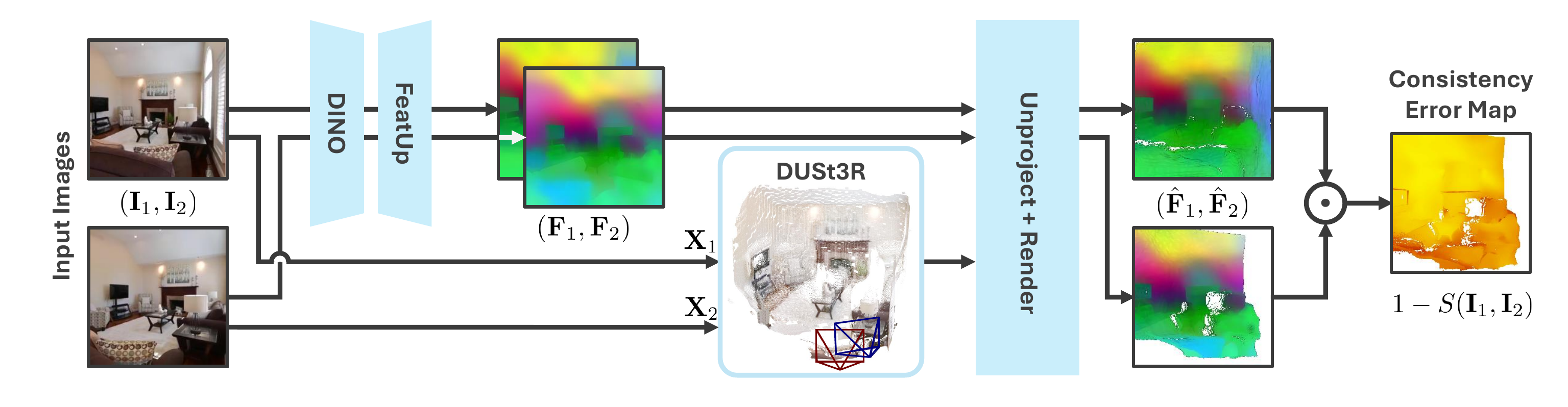}
    \caption{\textbf{Method overview.} Our metric evaluates the consistency between images $\mathbf{I}_1$ and $\mathbf{I}_2$. Given such a pair, we apply DUSt3R to obtain dense 3D point maps $\mathbf{X}_1$ and $\mathbf{X}_2$. These point maps are used to project upscaled DINO features $\mathbf{F}_1$, $\mathbf{F}_2$ into the coordinate frame of $\mathbf{I}_1$, via unprojecting and rendering. We compare the resulting feature maps $\hat{\mathbf{F}}_1$ and $\hat{\mathbf{F}}_2$ in pixel space to obtain similarity $S(\mathbf{I}_1,\mathbf{I}_2)$.
    }
    \label{fig:method}
\end{figure*}

\paragraph{Existing Metrics.}
Existing metrics used for quantifying image generation outputs include distribution-based metrics, such as the Fréchet Inception Distance (FID)~\citep{fid}, Kernel Inception Distance (KID)~\citep{binkowski2018demystifying}, Inception Score (IS)~\citep{salimans2016improved}, or the CLIP Maximum Mean Discrepancy (CMMD)~\citep{jayasumana2024rethinking}. While these metrics are used to measure the alignment of generated samples with a target distribution using pre-trained feature extractors, they do not measure 3D consistency, which is of utmost importance for multi-view generative models.
To this end, \citet{xie2024sv4d} proposed using the Fréchet Video Distance (FVD)~\cite{unterthiner2018towards} to measure the quality of generated sequences with moving camera.

To explicitly measure 3D consistency, \citet{watson2022novel} proposed to train a NeRF~\citep{nerf} from a subset of generated views and compare rendered novel views with the remaining generated set of images. This metric comes with several drawbacks, as it requires a large amount of generated images, does not work on sparsely observed scenes, is expensive to compute, and difficult to interpret: are dissimilarities between generated views and rendered novel views from the trained NeRF caused by inconsistencies in the multi-view generation pipeline or insufficient quality of the NeRF training?
As an alternative, \citet{Yu2023PhotoconsistentNVS} proposed TSED, a metric that checks whether image features detected in pairs of generated images respect the epipolar constraint, given the relative camera pose. As can be seen in Fig.~\ref{fig:existing}, it has certain limitations, e.g., it deems two images consistent when it finds enough matching features, ignoring obvious inconsistencies in the images. In contrast, MEt3R does not require camera poses as inputs, and we find that it is more aligned with perceptual assessment when looking at the results of individual methods.

%% file: sec/3_method.tex
\section{MEt3R: Measuring Consistency}

\label{sec:met3r}

In this section, we introduce MEt3R, our feed-forward metric to measure multi-view consistency.  Given two images as input, a metric for multi-view consistency should (1) penalize image pairs that are not consistent, and (2) must not penalize pairs that are consistent but deviate from a given ground truth or do not follow a desired distribution. Thus, we develop MEt3R to be orthogonal to image quality metrics, e.g., FID~\cite{fid}, and to pixel-wise reconstruction metrics, e.g., PSNR. 

An overview of MEt3R is shown in Fig.~\ref{fig:method}. Given two images $\mathbf{I}_1, \mathbf{I}_2$ as input, we first process them with DUSt3R to obtain dense 3D point maps for $\mathbf{I}_1$ and $\mathbf{I}_2$. Then, we obtain DINO~\cite{caron2021emerging} features on the original images and upscale them using FeatUp~\cite{fu2024featup}. We use the predicted point maps to unproject the upscaled features of both images into the 3D coordinate frame of $\mathbf{I}_1$ and render them separately onto the 2D image plane of the $1^{st}$ camera to obtain two projections. Lastly, we compute feature similarity on the projected features, leading to cosine similarity scores, which we denote as $S(\mathbf{I}_1, \mathbf{I}_2)$ and $S(\mathbf{I}_2, \mathbf{I}_1)$. 

\paragraph{MEt3R Definition.}
Given the scores $S(\mathbf{I}_1, \mathbf{I}_2)$ and $S(\mathbf{I}_2, \mathbf{I}_1)$, we can define MEt3R as,
\begin{equation}
\textnormal{MEt3R}(\mathbf{I}_1, \mathbf{I}_2) = 1 - \frac{1}{2} \Big(S(\mathbf{I}_1, \mathbf{I}_2) + S(\mathbf{I}_2, \mathbf{I}_1)\Big) \textnormal{,}
\end{equation}
which gives $\textnormal{MEt3R}(\cdot, \cdot) \in [0,2]$, lower is better, due to $S(\cdot,\cdot) \in [-1, 1]$, and is symmetric. We found $S$ to already behave approximately symmetric. Thus, in practice, $\textnormal{MEt3R}(\cdot, \cdot)$ can also be approximated well by only computing one direction of $S$ in case of runtime constraints.
We now provide the details for the DUSt3R reconstruction in Sec.~\ref{sec:DUSt3R} and feature similarity in Sec.~\ref{sec:features}

\subsection{Stereo Reconstruction with DUSt3R}
\label{sec:DUSt3R}
The core of our method relies on pose-free stereo reconstruction of pixel-aligned point clouds.
Given an image pair $\mathbf{I}_1$, $\mathbf{I}_2$, the DUSt3R~\cite{wang2023DUSt3R} model $\Psi$ regresses pixel-aligned 3D point clouds $\mathbf{X}_1 \in \mathbb{R}^{H \times W \times 3}$ and $\mathbf{X}_2 \in \mathbb{R}^{H \times W \times 3}$:
\begin{equation}
    \mathbf{X}_1, \mathbf{X}_2 = \Psi(\mathbf{I}_1, \mathbf{I}_2),
\end{equation}
where point locations of both, $\mathbf{X}_1$ and $\mathbf{X}_2$ are given in the camera space of $\mathbf{I}_1$. It does so by employing a shared ViT~\citep{dosovitskiy2020image} backbone to extract image features. Then, both feature maps are decoded by separate transformer decoders 
with cross-view attention that encodes a multi-view prior and shares important information between views. Finally the decoded features are regressed into point maps $\mathbf{X}_i$. For more details, please refer to the original work~\cite{wang2023DUSt3R}.

DUSt3R does not require camera poses, which is inherited by MEt3R. While MASt3R~\cite{Leroy2024MASt3r} additionally finds potentially useful feature correspondences between the two images, we do not make use of them in our method and hence stick with DUSt3R.

\subsection{High-Resolution Feature Similarity}
\label{sec:features}
Since both generated point maps contain points in the canonical coordinate frame of $\mathbf{I}_1$, we can use the point maps to project pixel-aligned features from camera space of $\mathbf{I}_2$ into that of $\mathbf{I}_1$. Instead of performing this projection and the subsequent comparison directly in RGB pixel space, we found it more suitable to perform them in feature space. The reason are view-dependent effects, such as different lighting, which often occurs in natural videos and negatively impacts RGB comparisons. We provide a detailed comparison between both approaches in Sec.~\ref{sec:ablation}. 

Concretely, we first use DINO~\cite{caron2021emerging} to obtain semantic features for $\mathbf{I}_1$ and $\mathbf{I}_2$. Then, since the corresponding feature maps are of low resolution and do not represent detailed structures, we upsample them using FeatUp \citep{fu2024featup}, which employs an image-adaptive upsampling, i.e., a stack of Joint Bilateral Upsamplers (JBUs) that learned to upsample low-resolution feature maps from DINO. It uses the high resolution image to transfer high frequency information to the upsampling process, allowing the upsampled features to faithfully reconstruct and preserve important details.

Let $\mathbf{F}_1$ and $\mathbf{F}_2$ denote the upsampled DINO features from images $\mathbf{I}_1$ and $\mathbf{I}_2$, respectively. We unproject both features into 3D space using the DUSt3R point maps and subsequently reproject them onto the camera frame of $\mathbf{I}_1$:
\begin{equation}
\hat{\mathbf{F}}_1 = \mathcal{P}(\mathbf{F}_1, \mathbf{X}_1) \textnormal{,} \quad\quad \hat{\mathbf{F}}_2 = \mathcal{P}(\mathbf{F}_2, \mathbf{X}_2) \textnormal{,}
\end{equation}
where $\mathcal{P}$ assigns each 3D point the feature vector from its corresponding pixel before rendering the feature point cloud using the PyTorch3D~\cite{ravi2020pytorch3d} point rasterizer.

Following the projections, we obtain $S(\mathbf{I}_1, \mathbf{I}_2)$ as the weighted sum of pixel-wise similarities between $\hat{\mathbf{F}}_1$ and $\hat{\mathbf{F}}_2$:
\begin{equation}
        S(\mathbf{I}_1, \mathbf{I}_2) = \frac{1}{|\mathbf{M}|} \sum^W_i \sum^H_j m^{ij}\frac{\hat{f}^{ij}_1 \cdot \hat{f}^{ij}_2}{\Vert\hat{f}^{ij}_1\Vert \Vert\hat{f}^{ij}_2\Vert} \textnormal{,}
        \label{eq:MEt3R}
    \end{equation}
where $m^{ij} := [\mathbf{M}]_{ij} $ is a boolean mask representing the overlapping region, $\hat{f}^{ij}_1 := [\hat{\mathbf{F}}_1]_{ij} $ and $\hat{f}^{ij}_2 := [\hat{\mathbf{F}}_2]_{ij}$.

\section{Multi-View Latent Diffusion Model}
\label{sec:mv_ldm}
Additionally to our metric, we provide an open-source multi-view latent diffusion model (MV-LDM). It is inspired by the architecture of CAT3D~\cite{gao2024cat3d}, which is not publicly available. While CAT3D is trained on top of proprietary image/video diffusion models, we initialize our model with Stable Diffusion 2.1~\cite{rombach2022high}. For a detailed description of MV-LDM, we refer to the appendix Sec.~\ref{sec:mv_ldm_details}. Our code and model are publicly available for further research.

\paragraph{Architecture and Training.}
MV-LDM encodes images into a latent space using a pre-trained VAE encoder from Stable Diffusion. Then, ray maps are concatenated to the input latents, providing camera pose information. We take the 2D UNet architecture, add attention between views at each UNet block, and finetune the full model on RealEstate10k for 1.65M iterations.  

\vspace{-0.2cm}
\paragraph{Anchored Generation.} We adopt the anchored generation strategy from CAT3D. When generating many views of a scene, the process starts by sampling four anchor images for widely distributed cameras, conditioned on a single input image. Then, in the second step, the remaining views are generated and conditioned on the closest anchor and the initial input image. The goal of the anchoring strategy is to prevent accumulating errors that often occur when generating target views autoregressively, conditioned on the previously generated views. When generating with anchors, the accumulation of errors can be effectively limited. We analyze the effect on consistency and image quality in Sec.~\ref{sec:comparison}.

%% file: sec/4_experiments.tex
\begin{figure*}
    \centering
    \includegraphics[width=\textwidth]{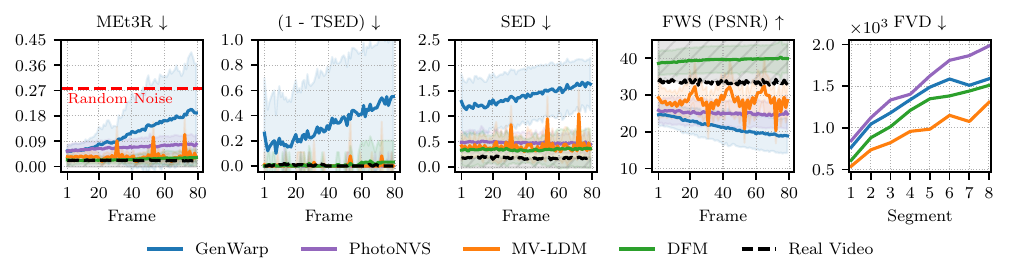}
    \caption{\textbf{Metric comparison.} We compare MEt3R against TSED, SED, FWS (PSNR), and FVD by computing average per-frame (/-segment for FVD) scores over many generated sequences. MEt3R can capture nuanced differences in the consistency of DFM, MV-LDM, and real videos, while TSED rates them all very similarly. Unlike MEt3R, SED does not capture increasing inconsistency for PhotoNVS and DFM. MEt3R also captures the influence of anchor views in MV-LDM (c.f. Sec.~\ref{sec:mv_ldm} and~\ref{sec:met3r_additional_details}) as structured high-frequency patterns. For MEt3R, the standard deviation gradually increases, starting from a small value. This behavior is expected due to the proximity of initial frames to the conditioning frame (c.f. Sec.~\ref{sec:obj_eval}) and is not the case for the other metrics. 
    }
    \label{fig:metric_comparison}
    \vspace{-0.2cm}
\end{figure*}

\section{Experiments}
\label{sec:formatting}

In this section, we evaluate MEt3R and existing generative models for multi-view and video generation. Specifically, we aim to answer the following questions:
\begin{itemize}
\itemindent=16pt
\item[\textbf{Q1:}] Does MEt3R fulfill the requirements for a useful consistency metrics as stated in Sec.~\ref{sec:method}, and how does it fare against previous metrics?
\item[\textbf{Q2:}] How consistent are the outputs of existing generative models for multi-view and video generation of objects and scenes?
\item[\textbf{Q3:}] How do individual design choices in MEt3R influence the metric quality?
\end{itemize}
We begin by introducing the experimental setup in Sec.~\ref{sec:experimental} before validating MEt3R (answering Q1) in Sec.~\ref{sec:validation}. Then, we address Q2 in Sec.~\ref{sec:comparison} and Q3 in Sec.~\ref{sec:ablation}.

\subsection{Experimental Setup}
\label{sec:experimental}
To evaluate MEt3R, we consider three sets of baselines for multi-view, video, and object-level generation models. In addition, we categorize the multi-view generation methods into three general classes: 1) single-view, 2) multi-view, and 3) 3D diffusion models.  

\paragraph{Multi-view Generation Models.} We consider GenWarp \cite{seo2024genwarp}, which is a single-view inpainting diffusion model, and PhotoNVS \cite{Yu2023PhotoconsistentNVS}, which is a two-view generation model that generates a single view at a time conditioned on the previous. Moreover, we consider DFM \cite{tewari2023forwarddiffusion}, which is a 3D diffusion method, and MV-LDM, our own open-source multi-view latent diffusion model, coupled with cross-view attention (c.f. Sec.~\ref{sec:mv_ldm} and~\ref{sec:mv_ldm_details}). For more details on baselines, we refer to Sec.~\ref{sec:baselines_details} in the appendix.

\begin{figure*}
    \centering
    \includegraphics[trim={0 0 0 0cm}, clip, width=\textwidth]{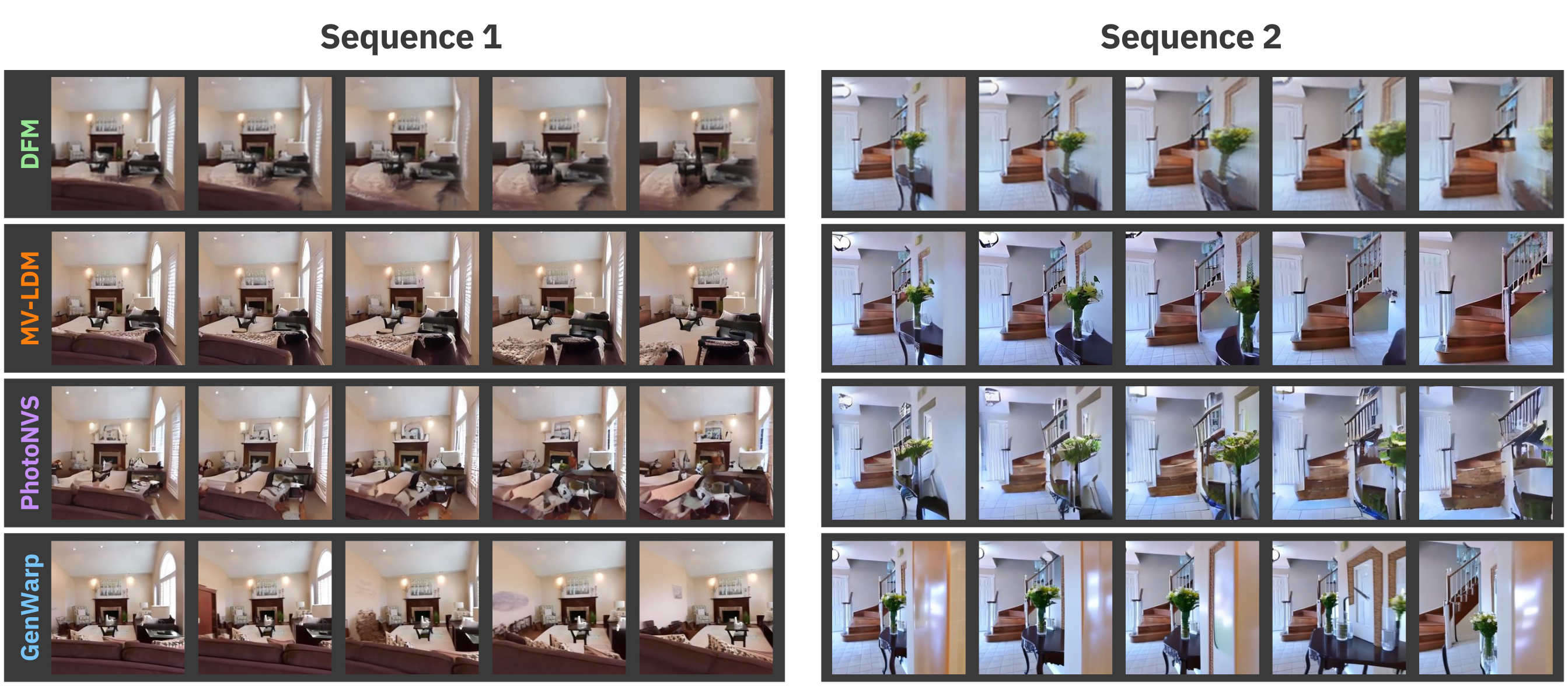}
    \caption{\textbf{Qualitative comparison of generated novel views.} We compare generated views of the multi-view generation method for the same conditioning view. We can extract certain characteristics: DFM is almost perfectly consistent but has lower image quality. PhotoNVS and MV-LDM are reasonably consistent on a structural scale but fail to produce consistent details. GenWarp fails to keep the structural consistency over the sequence while producing high-quality images. These observations are confirmed by MEt3R in Tab.~\ref{tab:metrics} and Fig.~\ref{fig:metric_comparison}. 
    }
    \label{fig:qualitative-comparison}
\end{figure*}


\paragraph{Video Generation Models.} We take Stable Video Diffusion (SVD)~\cite{blattmann2023stablevideodiffusionscaling}, Ruyi-Mini-7B~\cite{createai2024ruyi} and I2VGen-XL~\cite{zhang2023i2vgenxlhighqualityimagetovideosynthesis}, which are standard open source video diffusion models that can generate a full video from a single input image.

\paragraph{Object-Level Generation Models.} From object-level methods, we compare EpiDiff~\cite{huang2023epidiff}, SyncDreamer~\cite{syncdreamer}, and VideoMV~\cite{zuo2024videomv}. EpiDiff and SyncDreamer employ an underlying multi-view diffusion model, while VideoMV uses a video diffusion model to generate novel views of objects.


\paragraph{Dataset.} To faithfully benchmark with MEt3R, we collect 100 image sequences from the RealEstate10K~\cite{zhou2018stereo} test set. We take the first image for each sequence as the initial input, followed by 80 target poses, which the multi-view generation models generate. We perform consecutive pairwise evaluations on the generated images in a sliding-window fashion. In this way, we: 1) allow maximal projection area and more overlapping pixels to evaluate; 2) cover regions that are extrapolated and not visible in the input image; and 3) investigate the evolution of pairwise consistency as the camera pair moves further away from the input image. We set a standard resolution of $256^2$ as input to MEt3R. In the case of DFM, we upsample from $128^2$, and for GenWarp, we downsample from $512^2$ bilinearly. Similarly, we use identical test sequences for video diffusion models but limit them to 48 frames to address memory constraints. Note that we do not have explicit camera control over the generation and, therefore, are not equivalent in camera trajectories. The generated videos also differ in resolution, which we resize accordingly to the closest resolution of $256^2$ while preserving the aspect ratio. For object-level methods, we use Google Scanned Objects (GSO)~\cite{downs2022googlescannedobjectshighquality} dataset, which consists of $360^\circ$ views of objects. We subdivide the range $[0^\circ, 360^\circ]$ into 16 frames at $256^2$ resolution, forming a closed loop with the frame at $0^\circ$ as the conditioning. Both EpiDiff and SyncDreamer generate 16 frames, while VideoMV generates a fixed set of 32 frames, which we uniformly downsample to 16. 

\subsection{Validating MEt3R}
\label{sec:validation}
\paragraph{Computing 
lower bound
.}  We validate the efficacy of MEt3R by computing the lower bound that the baselines must follow. Intuitively, we can evaluate MEt3R on a dataset of real video sequences. Although they are assumed to be perfectly 3D consistent, a lower bound slightly above zero is observed, attributed to errors in point map alignment from DUSt3R~\cite{wang2023DUSt3R} and small 3D inconsistencies in DINO~\cite{caron2021emerging} features. The results for both multi-view generation baselines and real videos are shown in Fig.~\ref{fig:metric_comparison}.

\begin{table*}[t]
\footnotesize
\centering
  \subfloat[Quality vs. Consistency, Number of Parameters]{
\includegraphics[width=0.71\columnwidth]{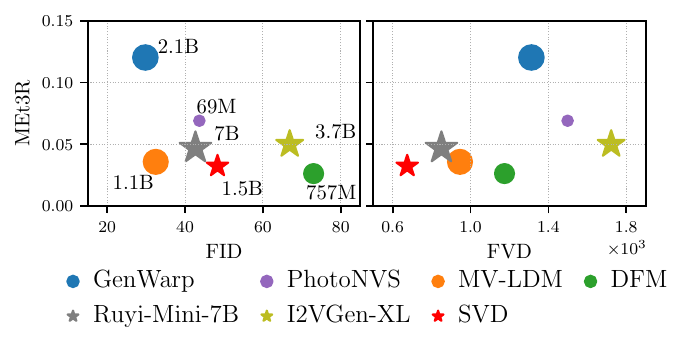}}
  \qquad
  \subfloat[Quantitative comparison with different metrics]{
\hspace{-0.5cm}

\begin{tabular}{l|cccccc}
\toprule[1pt]
\multicolumn{1}{c}{\textbf{Methods}} & \textbf{MEt3R} $\downarrow$ &   \textbf{TSED} $\uparrow$ &  \textbf{SED} $\downarrow$ &  \textbf{FVD} $\downarrow$ & \textbf{FID} $\downarrow$ & \textbf{FWS} \textbf{(PSNR)} $\uparrow$ \\  
\midrule
GenWarp ~\cite{seo2024genwarp} & 0.120  & 0.674 & 1.398 & 1312.7 & \textbf{29.80} & 21.41\\

PhotoNVS~\cite{Yu2023PhotoconsistentNVS} &  0.069 & \underline{0.996} & 0.479 & 1498.7 & 43.67 & 25.10 \\

MV-LDM (\textbf{Ours}) & \underline{0.036} & \textbf{0.998} & \underline{0.405} & \textbf{945.8} & \underline{37.29} & \underline{28.46} \\ 

DFM~\cite{tewari2023forwarddiffusion} & \textbf{0.026} & 0.990 & \textbf{0.346} & \underline{1174.6} & 73.02 & \textbf{39.56} \\

\midrule
I2VGen-XL~\cite{zhang2023i2vgenxlhighqualityimagetovideosynthesis} &  0.050 & - & - & 1722.6 & 66.88 & \underline{28.62} \\
Ruyi-Mini-7B~\cite{createai2024ruyi} & \underline{0.047} & - & - & \underline{850.5} & \textbf{42.67} & 28.01 \\
SVD~\cite{blattmann2023stablevideodiffusionscaling} & \textbf{0.032} & - & - & \textbf{674.6} & \underline{48.33} & \textbf{29.93} \\

\bottomrule[1pt]
\end{tabular}
  }

\caption{\textbf{Quantitative comparison.} Average MEt3R alongside TSED~\cite{Yu2023PhotoconsistentNVS}, SED~\cite{Yu2023PhotoconsistentNVS}, 
FVD~\citep{unterthiner2019fvd},
FID~\citep{fid}, and FWS (PSNR). (a) Plot comparing MEt3R with FID and FVD. (b) Quantitative comparison of multi-view and video generation baselines. Among multi-view methods, DFM achieves the best consistency in MEt3R, FWS (PSNR), and SED but the worst in FID. We attribute the low FID and high PSNR to blurry renderings, both of which are sensitive, whereas GenWarp delivers the best image quality with worse consistency but with a lower signal-to-noise ratio. In contrast, our MV-LDM achieves a favorable position in the image quality vs. consistency trade-off for multi-view generation. Unlike TSED and SED, MEt3R applies to generated video as it does not require camera poses. 
}

\label{tab:metrics}
\end{table*}
\paragraph{Comparison to other Metrics.}
We compare MEt3R with existing metrics to measure 3D consistency. As baselines, we consider SED~\cite{Yu2023PhotoconsistentNVS}, TSED~\cite{Yu2023PhotoconsistentNVS}, and FVD~\cite{unterthiner2018towards} for multi-view generation. We also compare with several variants of flow warping score (FWS) using RAFT~\cite{teed2020raftrecurrentallpairsfield} to warp one frame to another and compute PSNR, SSIM, LPIPS, and RMSE, among which we show the results for PSNR in Fig.~\ref{fig:metric_comparison}. For the remaining, we refer to Sec.~\ref{sec:flow_warp} in the appendix. In Fig.~\ref{fig:metric_comparison}, we plot per image-pair scores for all generated frames, averaged over 100 sequences. For FVD, we compare the distributions of image segments by splitting the sequences into chunks of 10 frames each. 
We find that MEt3R, SED, FWS (PSNR), and FVD increase as we progress through the image-pair sequence, suggesting a decrease in consistency, which is qualitatively visible in Fig.~\ref{fig:qualitative-comparison}. Although TSED captures this trend for GenWarp~\cite{seo2024genwarp}, it does not report a meaningful separation for other baselines. Unlike TSED and SED, MEt3R captures the gradual decrease in consistency for PhotoNVS~\cite{Yu2023PhotoconsistentNVS} and MV-LDM. For GenWarp, MEt3R captures this trend more accurately, starting with a lower score and standard deviation, as the first frame provides stronger conditioning for closer views with a larger overlap, resulting in better consistency. Furthermore, we observe sudden periodic spikes for MV-LDM in MEt3R, FWS (PSNR), and SED, attributed to transition artifacts when we switch between anchors during sampling (c.f. Sec.~\ref{sec:mv_ldm} and~\ref{sec:met3r_additional_details}). Unlike all other metrics, FVD cannot be applied to image pairs and requires a collection of frames. Ideally, a larger sample size is preferred to accurately capture and compare the underlying distribution of the generated and ground-truth image sequences~\cite{unterthiner2018towards}, to which FVD is sensitive. Moreover, both FVD and FWS (PSNR) are sensitive to blur. Specifically, DFM achieves worse FVD, which is supposed to be 3D consistent by design (c.f. Sec.~\ref{sec:experimental} and~\ref{sec:baselines_details}), while the real video, which is perfectly 3D consistent, gets worse than DFM in PSNR, as shown in Fig.~\ref{fig:metric_comparison}.

\subsection{Evaluations of Models}
\label{sec:comparison}
\subsubsection{Multi-View Generation}
Following the validation of MEt3R in comparison to other metrics, we now benchmark our multi-view generation baselines on the test sequences (c.f. Sec.~\ref{sec:experimental}). In \cref{tab:metrics}(a), we plot MEt3R against FID~\cite{fid} and FVD~\cite{unterthiner2019fvd} along with the respective model size in terms of the number of parameters. We find that GenWarp~\cite{seo2024genwarp} achieves the worst consistency in terms of MEt3R, where the contents of the scene change drastically as we transition from one image to another, which can be qualitatively observed in Figs.~\ref{fig:qualitative-comparison}, ~\ref{fig:qualitative_appendix_1} -~\ref{fig:qualitative_appendix_6}. This behavior is expected since GenWarp generates one image at a time. Meanwhile, PhotoNVS~\cite{Yu2023PhotoconsistentNVS} performs slightly better than GenWarp but produces low-quality results, which the FID captures quantitatively. GenWarp and PhotoNVS cannot learn an expressive multi-view prior since they have a single-sized context window, hindering their ability to produce consistent 3D results. 

Conversely, diffusing multiple views at a time induces a stronger prior towards 3D consistency, as in MV-LDM, where we see an overall improvement in MEt3R. Among all evaluated methods, MV-LDM achieves the best trade-off between 3D consistency and novel view quality, both qualitatively and quantitatively. Moving further towards 3D consistency, DFM~\cite{tewari2023forwarddiffusion} uses an underlying 3D representation and produces consistent novel views by design, which is captured quantitatively in the form of better MEt3R scores than MV-LDM. However, this strong inductive bias comes at the cost of blurry renderings pushing further from the ground-truth distribution, as reflected by the FID, and achieves a higher signal-to-noise ratio, as indicated by FWS (PSNR). This highlights that MEt3R only focuses on 3D consistency irrespective of image content and can, therefore, complement standard image quality metrics well.

\subsubsection{Video Generation}
A particular advantage of MEt3R is that it does not require camera poses, unlike TSED~\cite{Yu2023PhotoconsistentNVS} and SED~\cite{Yu2023PhotoconsistentNVS}, where it can be used directly on generated videos to measure consistency similar to FWS (PSNR). 

\begin{figure}[t]
    \centering
    \includegraphics[width=1.0\columnwidth]{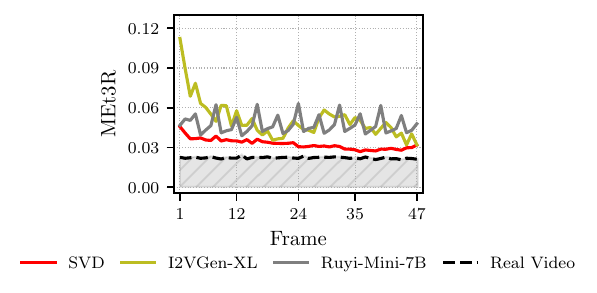}
    \caption{\textbf{Average pairwise MEt3R on generated videos.} Per-image-pair plot for MEt3R across 48 frames and averaged across 100 sequences of RealEstate10K~\cite{zhou2018stereo}. We find that SVD~\cite{blattmann2023stablevideodiffusionscaling} achieves the best MEt3R score, followed by Ruyi-Mini-7B~\cite{createai2024ruyi} and I2VGenXL~\cite{zhang2023i2vgenxlhighqualityimagetovideosynthesis}.}
    \label{fig:vldm_baseline}
\end{figure}
Table~\ref{tab:metrics} shows the average MEt3R, FID, FWS (PSNR), and FVD. Moreover, Fig.~\ref{fig:vldm_baseline} shows the average MEt3R per image pair for I2VGen-XL~\cite{zhang2023i2vgenxlhighqualityimagetovideosynthesis}, Ruyi-Mini-7B~\cite{createai2024ruyi} and SVD\cite{blattmann2023stablevideodiffusionscaling} which shows that SVD has better 3D consistency than Ruyi-Mini-7B and I2VGen-XL. However, SVD generates smoother and shorter camera trajectories, whereas Ruyi-Mini-7B and I2VGen-XL produce large motion at the expense of 3D consistency. For I2VGen-XL, as the inputs are out of distribution, MEt3R starts from a higher value followed by a gradual improvement as the model forces each progressing sample to be more in distribution while preserving similar global structures as in the initial input image. This behavior is also qualitatively visible in Figs.~\ref{fig:qualitative_appendix_1} -~\ref{fig:qualitative_appendix_6}. Meanwhile, Ruyi-Mini-7B shows several spikes indicating abrupt inconsistencies throughout the video sequence, which are attributed to unstable camera motion. Furthermore, in Tab.~\ref{tab:metrics}, FWS (PSNR) gives the lowest score to Ruyi-Mini-7B, indicating a slightly noisy generation even though it is more 3D consistent than I2VGen-XL.

\begin{figure}[t]
  \centering
  \includegraphics[width=\linewidth]{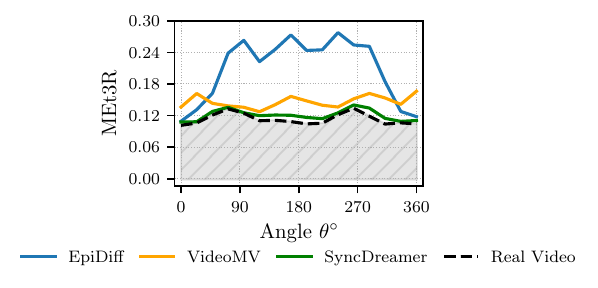}
   \caption{\textbf{Object-level evaluation on GSO~\cite{downs2022googlescannedobjectshighquality}.} Average pairwise MEt3R on 30 examples, each consisting of $360^\circ$ rotation around the object with loop closure, i.e., at $360^\circ$, we evaluate the first and the last frame. We find that SyncDreamer~\cite{syncdreamer} achieves the best in MEt3R, followed by VideoMV~\cite{zuo2024videomv} and EpiDiff~\cite{huang2023epidiff} while respecting the lower bound. }
   \label{fig:gso_eval}
\end{figure}
\subsubsection{Object-Level Generation}
\label{sec:obj_eval}
Lastly, we evaluate object-level diffusion models on GSO~\cite{downs2022googlescannedobjectshighquality} dataset. We consider EpiDiff~\cite{huang2023epidiff}, VideoMV~\cite{zuo2024videomv} and SyncDreamer~\cite{syncdreamer} as the baselines and are evaluated on $360^{\circ}$ camera rotation spread across 16 frames on 30 examples. We find that MEt3R can differentiate models with various levels of inconsistencies, which we report in Fig.~\ref{fig:gso_eval}. Although the baselines show good visual quality, consistency varies heavily. Notice how MEt3R captures the slightly increasing inconsistency for EpiDiff when moving further away from the condition, which suggests that the strength of conditioning plays an important role in producing better consistency (c.f. Fig.~\ref{fig:metric_comparison} and Sec.~\ref{sec:validation}). Further qualitative results are provided in Figs.~\ref{fig:qualitative_obj_1} -~\ref{fig:qualitative_obj_4} in the appendix.

\subsection{Analyzing Alternative Similarities}
\label{sec:alternative_sim}
\begin{figure}
    \centering
    \includegraphics[width=1.0\columnwidth]{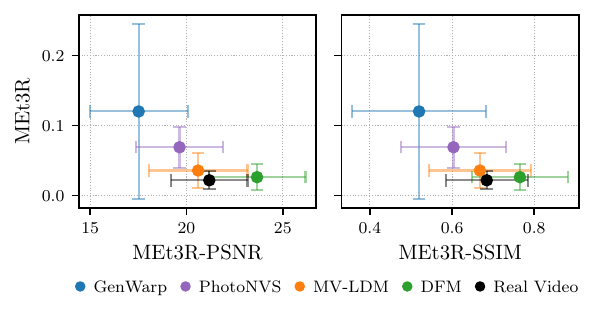}
    \caption{\textbf{Feature similarity ablation.} We compare MEt3R against versions of it that compare RGB projections via PSNR and SSIM. It can be seen that unlike MEt3R, PSNR and SSIM give better scores to DFM than to real videos. We attribute this to their sensitivity to view-dependent effects, such as lighting. Note that for real videos, the standard deviation of PSNR and SSIM are much higher, indicating a lower signal-to-noise ratio. 
    }
    \label{fig:psnr_MEt3R}
    \vspace{-0.3cm}
\end{figure}
We evaluate alternatives to the cosine similarity between DINO features as described in Sec.~\ref{sec:features}.
\vspace{-0.1cm}

\paragraph{Image Similarity.} Instead of projecting features onto a shared view, staying in RGB space would enable the use of classical image quality metrics such as PSNR and SSIM. Fig.~\ref{fig:psnr_MEt3R} provides a comparison of such variants $\mathrm{MEt3R_{PSNR}}$ and $\mathrm{MEt3R_{PSNR}}$ with MEt3R. While a reasonable negative correlation can be observed, DFM~\cite{tewari2023forwarddiffusion} outperforms the ground-truth video w.r.t. these metrics. We attribute this to the bias of PSNR and SSIM to blur, which is apparent in novel views generated by DFM due to its low resolution and reliance on pixelNeRF~\cite{pixelnerf} acting as an architectural bottleneck. In contrast, real videos exhibit view-dependent effects, including brightness variations and reflections, to which PSNR and SSIM are highly sensitive.
With MEt3R, we aim to abstract from these pixel-level inconsistencies and instead provide a metric that robustly measures the 3D consistency of generative approaches. Therefore, we opt for similarities in a suitable feature space.

\begin{figure}
    \centering
    \includegraphics[width=1.0\columnwidth]{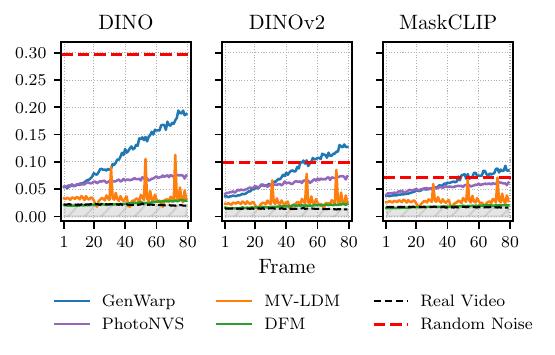}
    \caption{\textbf{Feature backbone ablation.} We analyze the effect of different feature backbones on MEt3R. While DINOv2~\cite{oquab2023dinov2} and MaskCLIP~\cite{ding2023maskclip} can be employed as well, we found DINO~\cite{caron2021emerging} features to lead to a more informative separation of models.}
    \label{fig:ablation-backbones}
    \vspace{-0.3cm}

\end{figure}

\vspace{-0.2cm}
\paragraph{Feature Backbones.}
In Fig.~\ref{fig:ablation-backbones}, we evaluate MEt3R in combination with DINOv2~\cite{oquab2023dinov2} and MaskCLIP~\cite{ding2023maskclip} as alternatives to DINO~\cite{caron2021emerging} in the feature backbone. DINOv2 and MaskCLIP strongly compress the values in a tighter range, reducing the gap between extremely inconsistent and consistent generation. We find that DINO features provide a better separation of model performance and capture substantial inconsistencies more reliably, as seen from the random noise. Nevertheless, MEt3R is flexible with this design choice as better and more 3D consistent feature backbones can improve and reduce the lower bound further. 
\label{sec:ablation}

\section{Conclusion} \label{sec:conclusion}
We presented MEt3R, a novel metric for 3D consistency of generated multi-view images. Given the huge success of large-scale image diffusion models and their applications as strong priors for the generation of multi-view images as a form of 3D representation, purely distribution-based metrics like FVD are insufficient to properly evaluate the 3D capabilites of such methods.
First, MEt3R leverages DUSt3R to warp images robustly into a shared view without relying on ground truth camera poses as input.
Secondly, by computing similarities in the feature space of DINO, MEt3R abstracts from view-dependent effects.
As a result, we show that our proposed metric can be effectively employed for comparing the performance of multi-view generation approaches like our open-source multi-view latent diffusion model, which finds the best trade-off between novel view quality and consistency.
Given the recent trend towards large video models, we see great potential for MEt3R to effectively evaluate their 3D consistency since no ground truth camera poses are required.
 
\section*{Acknowledgements}
This project was partially funded by the Saarland/Intel Joint Program on the Future of Graphics and Media.
Thomas Wimmer is supported through the Max Planck ETH Center for Learning Systems.

%% file: sec/X_suppl.tex
\clearpage
\begin{appendices}
\maketitlesupplementary

The supplementary materials are structured as follows. First, 
we provide detailed information about our multi-view latent diffusion model (MV-LDM) in Sec.~\ref{sec:mv_ldm_details}. Then, we compare MEt3R with the FWS variants in Sec.~\ref{sec:flow_warp}, followed by a discussion on MEt3R metric in Sec.~\ref{sec:MEt3R_details}. Finally, we present additional details on the multi-view generation baselines in Sec.~\ref{sec:baselines_details} and their corresponding runtime statistics in Sec.~\ref{sec:runtime}. Please also note our \textbf{supplementary video}, showcasing evaluations in motion.

\section{Multi-View Latent Diffusion Model}
\label{sec:mv_ldm_details}
This section presents further details for MV-LDM, including the architectural components, training, and sampling details. 

\subsection{Architecture.} 
\begin{figure*}[t]
    \centering
    \includegraphics[width=1\linewidth]{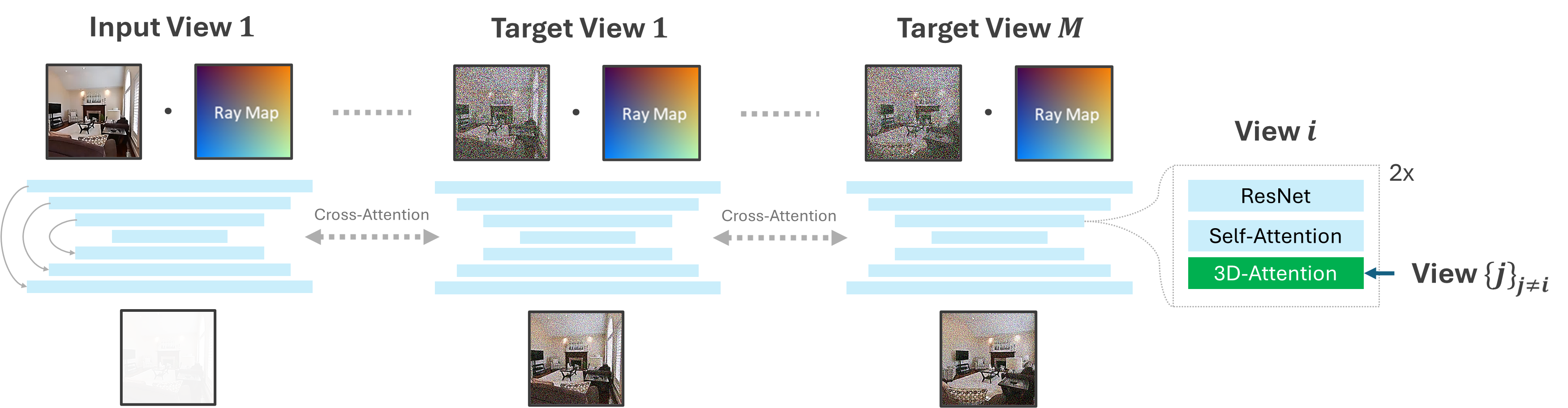}
    \caption{\textbf{MV-LDM. } Architecture overview of MV-LDM, which consists of a shared 2D UNet initialized from Stable Diffusion 2.1 \cite{rombach2022high} across multiple input views with cross-view attentions (3D attention) in between for modeling multi-view prior.}
    \label{fig:mvldm_arch}
\end{figure*}

Like CAT3D \cite{gao2024cat3d}, our architecture is based on a multi-view 2D UNet shared across multiple input views with 3D self-attention at each UNet block. We initialize the UNet weights with Stable Diffusion 2.1 \cite{rombach2022high} and replace each attention layer with a 3D self-attention layer from MVDream \cite{shimvdream} where each token from one view attends to all tokens from the other views. This accounts for 1.1B parameters for the multi-view UNet and 83.7M for the VAE. Due to memory and resource limitations, we fix the total number of concurrent views to 5, including the target and the conditioning views. Figure~\ref{fig:mvldm_arch} shows the architecture of MV-LDM. We apply a VAE encoder and map the input images $(H \times W \times 3)$ into latent representation $(\frac{H}{8} \times \frac{W}{8} \times 4)$. For the low-resolution latent maps, we generate the ray encodings of shape $(\frac{H}{8} \times \frac{W}{8} \times 6)$, which consists of a 3-dimensional origin and a 3-dimensional direction vector in relative camera space and concatenate it along the channel dimension. 

\subsection{Training and Evaluation with MEt3R} 
\label{sec:met3r_additional_details}
\paragraph{Dataset.} We use RealEstate10K~\cite{zhou2018stereo}, which consists of 80K video sequences accounting for 10 million frames. During training, we randomly select a video sequence and the corresponding conditioning and target views that satisfy the following criteria:
\begin{itemize}
    \item Sample 2 conditioning views (left and right) at frame number $f_L$ and $f_R$ with frame distance $d_c = f_R - f_L$ satisfying $50 \leq d_c \leq 180$.
    \item Sample 3 target views with distance $d_t$ from the conditioning view that satisfies $ f_L - 100 \leq d_t \leq f_R + 100$.
\end{itemize}
Afterward, we transform the absolute poses into relative poses with respect to the first conditioning view.

\paragraph{Training.} The training procedure follows DDPM~\cite{ddpm}, sampling a noise level $t$, applying that to all given latent images and training the network to predict the noise present in the image. We randomly select the conditioning views $N$ between 1 or 2 and the target views $M$ between 3 and 4, respectively, to allow for single and few-view novel view generation. 
We linearly vary the beta schedule from 0.0001 to 0.02 for the forward diffusion process and train MV-LDM for a total of 1.65M iterations with an effective batch size of 24 at resolution $256^2$. We use AdamW~\cite{loshchilov2019decoupledweightdecayregularization} optimizer with a constant learning rate of $2e^{-5}$. During sampling, the network can receive a combination of existing and pure noise images with camera ray encodings to perform conditional generation. The backward diffusion process is done with $\epsilon$-parameterization defined as the output of the model $\boldsymbol{\epsilon}_{\theta}$:
\begin{equation}
    \boldsymbol{\epsilon}_{pred} = \boldsymbol{\epsilon}_{\theta}\left(\mathbf{z}_t,\mathbf{c}_t, t\right) \textit{,}
\end{equation}
where $\boldsymbol{\epsilon}_{pred} = (\boldsymbol{\epsilon}^i_{pred})^{M}_{i=1}$ is the predicted noise latent,  $\mathbf{z}_t = (\mathbf{z}^i_t)^{M}_{i=1}$ is the noisy latent, $\mathbf{c}_t = (\mathbf{c}^j_t)^{N}_{j=1}$ is the clean latent at the timestep $t$, whereas $M$ and $N$ are the number of target and conditioning views, respectively. The predicted noise is used to make a step in the direction of a sample in the target distribution under the DDIM~\cite{ddim} formulation. For classifier-free guidance, we randomly drop the clean conditioning views with a probability of 10\%, and during sampling, we apply a guidance scale of 3 similar to CAT3D~\cite{gao2024cat3d}. 

For training, we apply the standard diffusion loss on the predicted mean noise as the mean-squared error (MSE) against the ground truth noise:
\begin{equation}
    \mathcal{L} = ||\boldsymbol{\epsilon} - \boldsymbol{\epsilon}_{\theta}\left(\mathbf{z}_t,\mathbf{c}_t, t\right)||^2_2 \textit{,}
\end{equation}
where $\boldsymbol{\epsilon} = (\boldsymbol{\epsilon}^i)^{M}_{i=1}$ and $\boldsymbol{\epsilon}^i \sim \mathcal{N}(\mathbf{0}, \mathbf{I})$ is the ground truth noise for each target view.

\begin{figure}[t]
    \centering
    \includegraphics[width=.8\columnwidth]{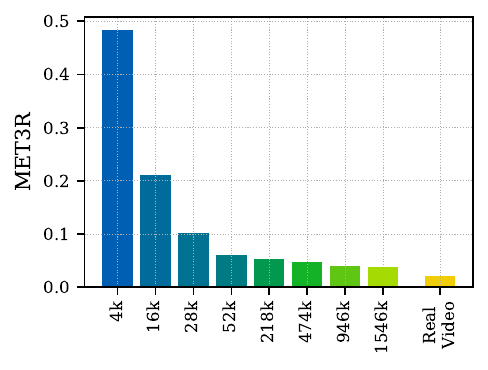}
    \caption{\textbf{MEt3R at different training iterations.} As we continue to train MV-LDM, we see a consistent improvement in 3D consistency, which is an expected behavior. Furthermore, in the beginning, the improvements are large, which slows down and saturates in the later iterations.}
    \label{fig:training_evolution}
\end{figure}

\paragraph{Training evolution of MEt3R.} Figure~\ref{fig:training_evolution} shows the trend in 3D consistency in terms of MEt3R over training iterations, showing consistent improvements with longer training. There is a significant improvement in the initial 100k, and afterward, it saturates near 1M iterations. 

\begin{figure}[t]
    \centering
    \includegraphics[width=1\columnwidth]{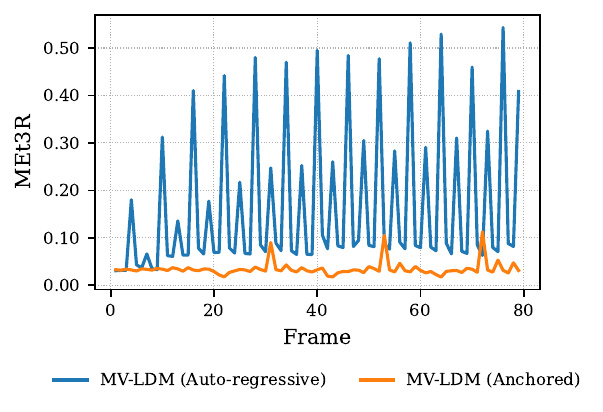}
    \caption{\textbf{Anchored vs. autoregressive.} Per-image-pair MEt3R on 2 different sampling strategies. For autoregressive sampling, we see significant and periodic spikes becoming larger as we progress and show the effect of compounding error, i.e., sequentially generating new frames and anchors conditioned on the previously generated ones. As illustrated in Fig.~\ref{fig:anchor_illust}, autoregressive sampling produces several anchor-to-anchor transitions causing these periodic spikes. On the other hand, anchored generation limits the effect of compounding error by generating all anchors in parallel.}
    \label{fig:anchored_vs_sequential}
\end{figure}

\begin{figure*}[t]
    \centering
    \includegraphics[width=1\linewidth]{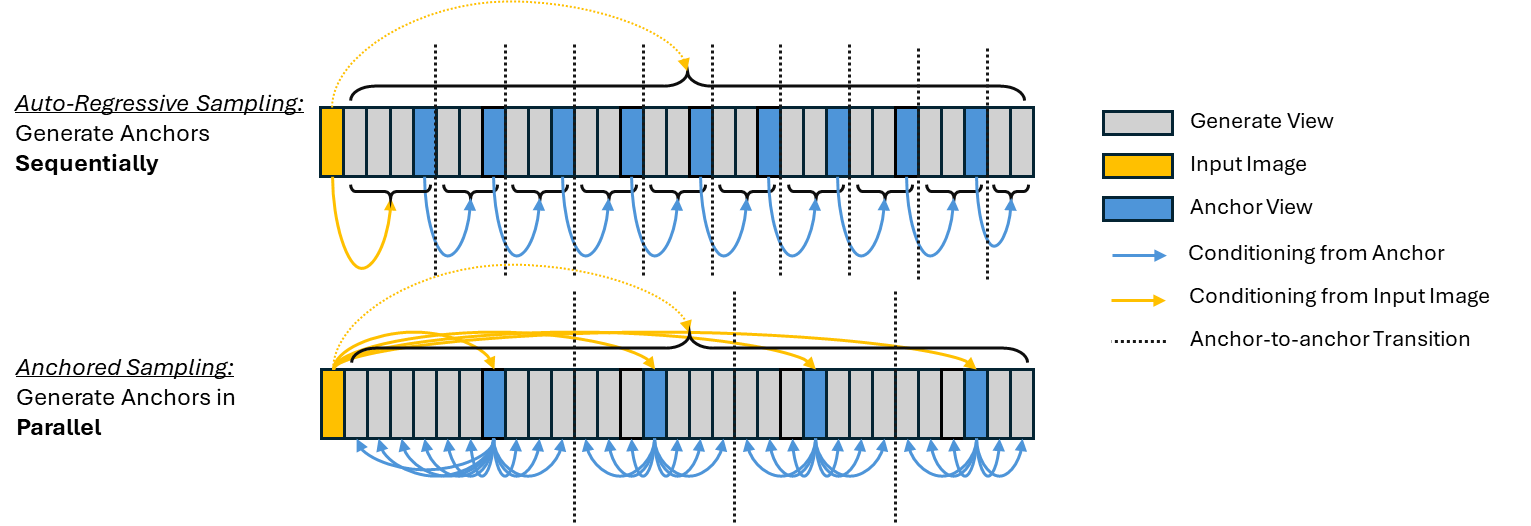}
    \caption{\textbf{Anchored vs. Autoregressive sampling schemes.} An illustration of the differences in the sampling schemes. In autoregressive sampling, we start from the initial input image and generate a set of target frames. The next set of frames is conditioned both on the input image and on the last frame (anchor) of the previously generated set. With this sampling strategy, we see several anchor-to-anchor transitions and results in large inconsistencies as visible in Fig.~\ref{fig:anchored_vs_sequential}. Whereas using anchored generation i.e., generate anchors first and then sample the remaining conditioned on the closest anchor and the input image. With this strategy, we observe significantly fewer anchor-to-anchor transitions, limited error accumulation, and relatively stable and lower MEt3R across the image pairs.}
    \label{fig:anchor_illust}
\end{figure*}

\paragraph{Anchored vs. autoregressive sampling.} We further test MEt3R with two sampling strategies, i.e., (1) autoregressively generating new target views and new anchors, conditioned on the previous anchor, and (2) using anchored sampling where we generate anchors first and then the rest as described in Sec.~\ref{sec:mv_ldm}. Fig.~\ref{fig:anchored_vs_sequential} shows the average MEt3R plot per image-pair, showing the improvements with anchored sampling. We observe many diverging peaks attributed to several anchor-to-anchor transitions and accumulating errors for autoregressive sampling. For anchored sampling, the anchors are generated together first, followed by generating the rest. This limits the error accumulation and allows for fewer anchor-to-anchor transitions. We refer to Fig.~\ref{fig:anchor_illust} for a visual illustration of anchored and autoregressive sampling schemes.

\paragraph{MEt3R on multiple scales.} 
In Tab.~\ref{tab:multi_scale}, we investigate the effect of image resolutions on MEt3R compared to SED~\cite{Yu2023PhotoconsistentNVS}. We find that SED is highly sensitive to variation in image resolution with a significant increase at $128^2$. This is expected since SED computes the geometric distance of each correspondence from their epipolar line in the 2D-pixel space. Meanwhile, MEt3R is more robust, attributed to the measurement in the feature space (c.f. Sec.~\ref{sec:met3r}), thus maintaining only minor differences in the scores. Although the differences are small, we still recommend using a similar resolution for all baselines for a fair comparison.
\begin{table}[t]
\centering

\resizebox{\columnwidth}{!}{\begin{tabular}{l|cc|cc|cc}
\toprule[1pt]
 & \multicolumn{2}{c|}{$\mathbf{256^2}$}  &  \multicolumn{2}{c|}{$\mathbf{224^2}$}  &  \multicolumn{2}{c}{$\mathbf{128^2}$}  \\
 & MEt3R & SED & $\textit{Diff}_\textit{MEt3R}$ & $\textit{Diff}_\textit{SED}$ & $\textit{Diff}_\textit{MEt3R}$ & $\textit{Diff}_\textit{SED}$  \\
\midrule

GenWarp~\cite{seo2024genwarp} & 0.120 & 1.398 & \cellcolor{green!25}-$2.58\%$  & \cellcolor{red!25}-$74.84\%$ & \cellcolor{green!25}+$6.61\%$ & \cellcolor{red!25}-$80.37\%$  \\

PhotoNVS~\cite{Yu2023PhotoconsistentNVS} & 0.069 & 0.479 & \cellcolor{green!25}-$3.89\%$  & \cellcolor{red!25}-$26.23\%$ & \cellcolor{green!25}+$1.60\%$ & \cellcolor{red!25}-$42.50\%$  \\

MV-LDM (\textbf{Ours}) & \underline{0.036} & \underline{0.405} & \cellcolor{red!25}-$3.88\%$  & \cellcolor{green!25}+$2.29\%$ & \cellcolor{green!25}+$16.66\%$ & \cellcolor{red!25}+$34.23\%$  \\

DFM~\cite{tewari2023forwarddiffusion} & \textbf{0.026} & \textbf{0.346} & \cellcolor{green!25}-$9.35\%$  & \cellcolor{red!25}-$52.84\%$ & \cellcolor{green!25}+$5.73\%$ & \cellcolor{red!25}-$70.20\%$  \\
\midrule
Real Video & 0.022 & 0.181 & \cellcolor{green!25}-$2.46\%$ & \cellcolor{red!25}+$18.99\%$ & \cellcolor{green!25}+$47.87\%$  & \cellcolor{red!25}+$148.17\%$ \\

\bottomrule[1pt]
\end{tabular}}
\caption{\textbf{MET3R vs. SED on multiple resolutions}. 
We show differences in SED~\cite{Yu2023PhotoconsistentNVS} and MEt3R for the baseline multi-view generation models over changing image resolution against the base resolution of $256^2$ in percentage. MEt3R is more robust to variations in the input resolution since it measures in feature space, unlike SED, which measures in pixel space (c.f. Sec.~\ref{sec:met3r}). Here, SEDs total scale is less than one order of magnitude larger than MEt3Rs, while its variations are more than one order of magnitude larger in most cases.}
\label{tab:multi_scale}
\end{table}
\section{Comparison of FWS Variants}
\label{sec:flow_warp}

Other metrics based on flow warping score (FWS) have been introduced to measure consistency, which uses optical flow, e.g., RAFT~\cite{teed2020raftrecurrentallpairsfield}. Given a pair of images, it first computes optical flow, which is used to warp one image into the other. Then, metrics such as SSIM, LPIPS, PSNR, and RMSE are computed to quantify multi-view consistency. 

\begin{figure}[t]
    \centering
    \includegraphics[width=1\columnwidth]{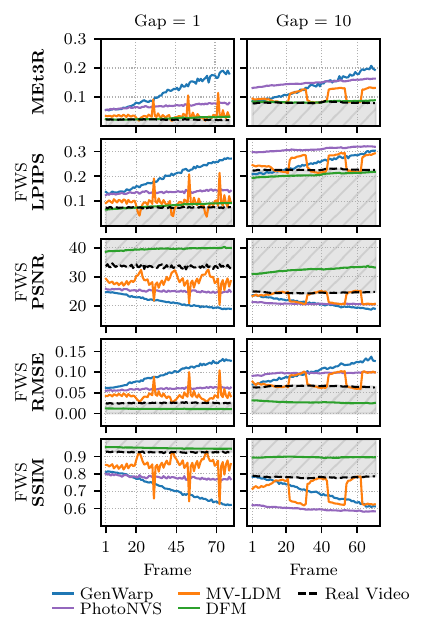}
    \caption{\textbf{MEt3R comparison to existing Flow Warping Score (FWS).} For a gap (frame distance) of 10 between the image pair, MET3R is more robust and does not violate the lower bound, unlike FWS.}
    \label{fig:gaps}
\end{figure}

In Tab.~\ref{tab:flow_warping_scores}, we evaluate both multi-view and video generation baselines on FWS and MEt3R. We find that most variants, including PSNR, SSIM, and RMSE, rank DFM better than real video among multi-view generation methods because of their sensitivity to blur, pixel-level perturbation, and noise (c.f. Sec.~\ref{sec:comparison} and~\ref{sec:alternative_sim}). Meanwhile, MEt3R and FWS (LPIPS) ignore such perturbations and rely on feature and perceptual similarity, respectively. However, as shown in Fig.~\ref{fig:gaps}, FWS (LPIPS) suffers at higher frame distances between input pairs, where DFM, GenWarp, and MV-LDM can score better than real videos.  
\begin{table}[t]
\centering
\resizebox{\columnwidth}{!}{%
\begin{tabular}{l|c|cccc}
\toprule[1pt]
 &\multicolumn{1}{c|}{}  & \multicolumn{4}{c}{\textbf{FWS}} \\
 &  \textbf{MEt3R} $\downarrow$ & \textbf{PSNR} $\uparrow$ & \textbf{SSIM} $\uparrow$ & \textbf{LPIPS} $\downarrow$ & \textbf{RMSE} $\downarrow$ \\
\midrule

GenWarp ~\cite{seo2024genwarp} & 0.120  & 21.41 & 0.716 & 0.200 & 0.097\\

PhotoNVS~\cite{Yu2023PhotoconsistentNVS} &  0.069 & 25.10 & 0.779 & 0.137 & 0.060 \\

MV-LDM (\textbf{Ours}) & 0.036 & 28.46 & 0.851 & 0.095 & 0.044 \\ 

DFM~\cite{tewari2023forwarddiffusion} & \underline{0.026} & \textbf{39.56} & \textbf{0.948} & \underline{0.082} & \textbf{0.011}  \\

Real Video & \textbf{0.022} & \underline{33.52} & \underline{0.924} & \textbf{0.075} & \underline{0.026}  \\

\midrule

I2VGen-XL~\cite{zhang2023i2vgenxlhighqualityimagetovideosynthesis} &  0.050 & 28.62 & 0.844 & 0.107 & 0.044 \\

Ruyi-Mini-7B~\cite{createai2024ruyi} & 0.047 & 28.01 & 0.831 & 0.133 & 0.043 \\

SVD~\cite{blattmann2023stablevideodiffusionscaling} & \underline{0.032} & \underline{29.93} & \underline{0.890} & \underline{0.079} & \underline{0.038} \\

Real Video & \textbf{0.022} & \textbf{33.60} & \textbf{0.925} & \textbf{0.074} & \textbf{0.026} \\

\bottomrule[1pt]
\end{tabular}}
\caption{\textbf{Comparison of flow warping scores (FWS) with MEt3R}. We report the results on multi-view and video generation methods, with FWS variants including PSNR, SSIM, LPIPS, and RMSE. }
\label{tab:flow_warping_scores}
\end{table}

\section{Additional MEt3R Architectural Details}
\label{sec:MEt3R_details}
\begin{figure}[t]
    \centering
    \includegraphics[width=1\columnwidth]{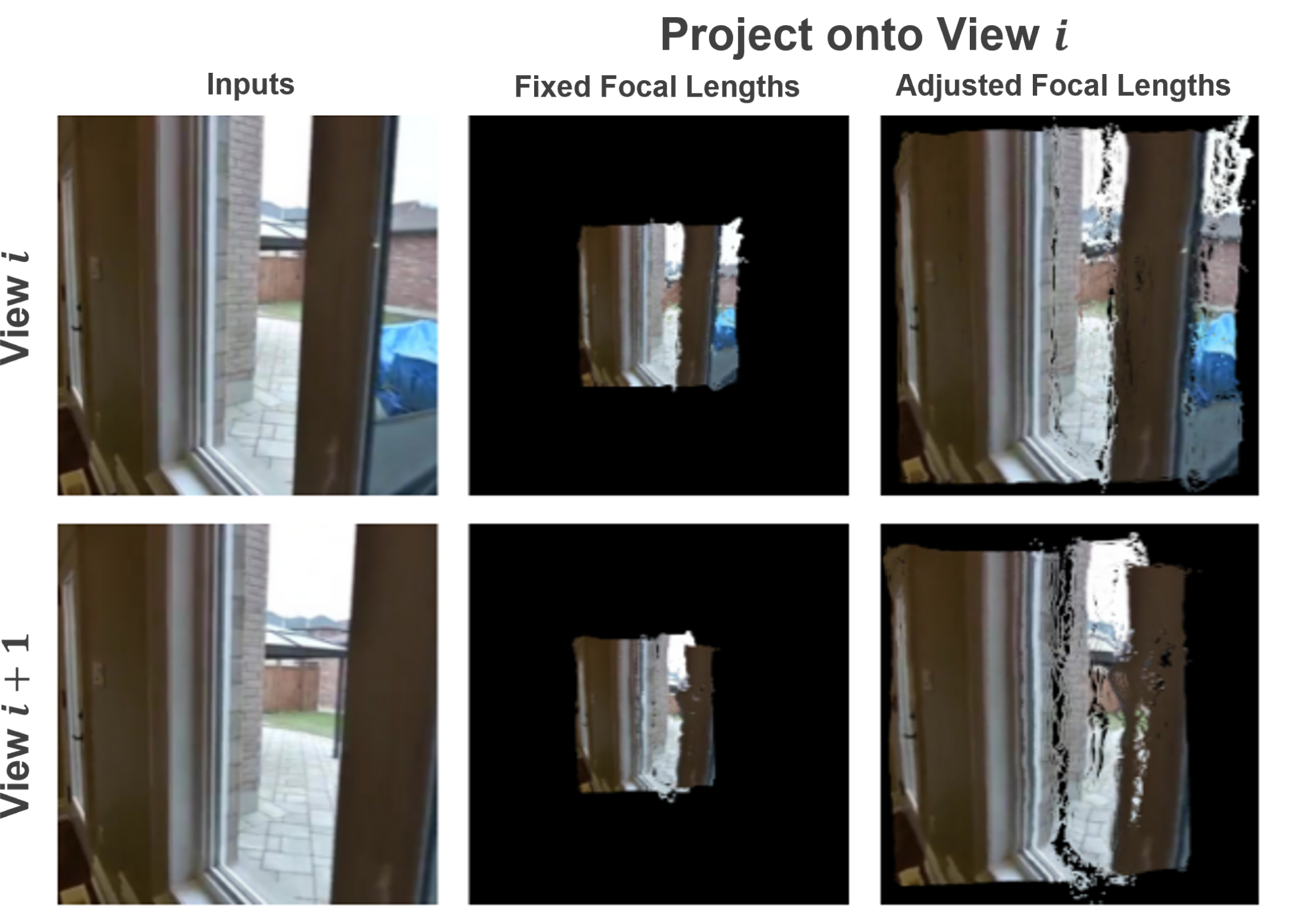}
    \caption{\textbf{Fixed vs. adjusted projection matrix.} With fixed focal length, the projection area varies across different scales of DUSt3R~\cite{wang2023DUSt3R} point maps. We automatically adjust the focal length for each example pair to allow maximal projection and, therefore, more pixels for evaluating feature similarity.}
    \label{fig:projection}
\end{figure}

This section presents additional details on the MEt3R pipeline, including the projection of both point maps to the first view and a description of the overlap mask used.
\paragraph{Projection matrix.} Figure~\ref{fig:projection} shows a side-by-side comparison of different projections we obtain using 1): fixed focal length and 2): Adjusting focal length based on the scale of canonical point map. We compute the canonical point map $\mathbf{X}_{canon}$ as the weighted sum of the point maps pair $\mathbf{X}_{i}$ and $\mathbf{X}_{i+1}$ using their corresponding confidences $\mathbf{C}_{i}$ and $\mathbf{C}_{i+1}$ from DUSt3R~\cite{wang2023DUSt3R} as,
\begin{equation}
    \mathbf{X}_{canon} = \frac{\mathbf{C}_{i} \odot \mathbf{X}_{i} + \mathbf{C}_{i+1} \odot \mathbf{X}_{i+1}}{\mathbf{C}_{i} + \mathbf{C}_{i+1}}
\end{equation}

Then, we extract the $x$, $y$, and $z$ coordinate maps from $\mathbf{X}_{canon}$ as $\mathbf{X}, \mathbf{Y}, \mathbf{Z} \in \mathbb{R}^{H \times W}$. Moreover, DUSt3R already implements this in their codebase, which we incorporate in MEt3R as shown in Alg.~\ref{algo:focal_adj}. The computed focal length $f_x$ and $f_y$, along with the principal point offsets $c_x$ and $c_y$, are used to form the projection matrix.

\algnewcommand\algorithmicinput{\textbf{Input:}}
\algnewcommand\algorithmicoutput{\textbf{Output:}}
\algnewcommand\Input{\item[\algorithmicinput]}
\algnewcommand\Output{\item[\algorithmicoutput]}

\begin{algorithm}
    \begin{algorithmic}[1]
        \Input {2D pixel position $\mathbf{U}, \mathbf{V} \in \mathbb{R}^{H \times W}$, 
        3D position $\mathbf{X}, \mathbf{Y}, \mathbf{Z} \in \mathbb{R}^{H \times W}$} 
        
        \Output{$f_x, f_y$}

        \State {$\mathbf{Q}_x = \frac{\mathbf{U}\odot\mathbf{Z}} {\mathbf{X}}$} \Comment{$\odot$ is the Hadamard Product}
        \State {$\mathbf{Q}_y = \frac{\mathbf{V}\odot\mathbf{Z}} {\mathbf{Y}}$}

        \State {$f_x = median(\mathbf{Q}_x)$} \Comment{Across spatial dimension}
        \State {$f_y = median(\mathbf{Q}_y)$}
        
    \end{algorithmic}
    \caption{Computing focal length given 2D grid of pixel positions and 3D canonical point maps}
     \label{algo:focal_adj}
\end{algorithm}

\begin{figure}[t]
    \centering
    \includegraphics[width=1\columnwidth]{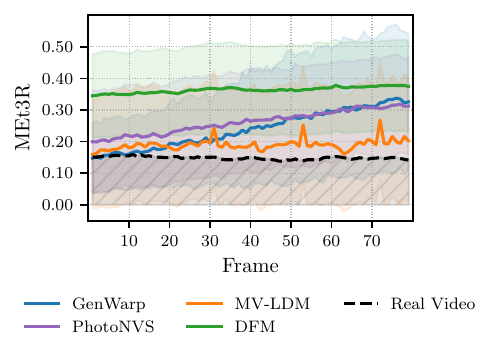}
    \caption{\textbf{MEt3R without overlap mask.} Per-image-pair MEt3R without normalizing against the overlap mask. Under this setting, DFM~\cite{tewari2023forwarddiffusion} is worse than all other baselines in 3D consistency, even though it has a strong inductive bias, which forces its results to be 3D consistent at the expense of blur. Whereas PhotoNVS~\cite{Yu2023PhotoconsistentNVS} and GenWarp~\cite{seo2024genwarp} are similar, both of which increase gradually, whereas MV-LDM stays relatively low with visible spikes due to anchor-to-anchor transitions.}
    \label{fig:MEt3R_nomask}
\end{figure}

\paragraph{Overlap mask.} We normalize MEt3R with an overlap mask $\mathbf{M}$ as formulated in Eq.~\ref{eq:MEt3R} which is a crucial step. During rasterization, we set the background values to a large negative value $\eta$ for each channel and subsequently build the mask using the background values for each projected view, i.e,
\begin{equation}
    m^k_{ij} = \begin{cases}
      0 & \text{\textbf{if} }  p^k_{ij} = \eta  \\
      1 & \text{\textbf{otherwise}}
    \end{cases}
\end{equation}
where $m^k_{ij} = [\mathbf{M}^k]_{ij}$ is the mask for $k^{th}$ view, $p^k_{ij} = [\mathbf{P}^k]_{ij}$ are the pixel values after projection and rasterization. We set $\eta = -10000$, and we perform pixel-wise multiplication of both masks $\mathbf{M}^{i}$ and $\mathbf{M}^{i+1}$ to get the overlap mask $\mathbf{M}$:
\begin{equation}
    \mathbf{M} = \mathbf{M}^{i} \odot \mathbf{M}^{i+1} 
\end{equation}
Figure.~\ref{fig:MEt3R_nomask} shows MEt3R without normalizing against the overlap mask $\mathbf{M}$ in Eq.~\ref{eq:MEt3R}. Instead, we take the average of the similarity scores for all pixels. Compared to MEt3R (c.f. Fig.~\ref{fig:metric_comparison}),  the lower bound gets significantly larger with a large offset, while DFM~\cite{tewari2023forwarddiffusion} gets worse than all other baselines. Meanwhile, PhotoNVS~\cite{Yu2023PhotoconsistentNVS} gets almost similar to GenWarp~\cite{seo2024genwarp}. This contradicts both the theoretical expectation and the visual judgment about the 3D consistency of the baselines. In addition, the standard deviations for all baselines are large and correspond to noisy scores for individual image pairs across the test sequences. However, some key features, such as spikes from anchor-to-anchor transitions in MV-LDM and the gradual increase in MEt3R due to decreasing 3D consistency, are still visible.

\section{Additional Details on Multi-View Generation Models}
\label{sec:baselines_details}
In the following, we present additional details on the multi-view generation baselines.  

\paragraph{GenWarp.} GenWarp~\cite{seo2024genwarp} employs a two-step approach, i.e., project and in-paint. With a monocular depth estimator, it predicts depth maps for the input image and un-projects the RGB in 3D space. The 3D points are rendered onto a target view, followed by inpainting with an image-to-image diffusion model. GenWarp generates only one view at a time. For every novel view, we condition the model on the fixed input view for every novel view, as an autoregressive approach diverges very quickly due to error accumulation.  

\paragraph{PhotoNVS.} Just like GenWarp, PhotoNVS~\cite{Yu2023PhotoconsistentNVS} also generates a single view at a time given a conditioning image. However, by employing a score-based diffusion UNet architecture for both views with cross-view attention in-between, it can always condition on the last generated frame in an autoregressive fashion, improving multi-view consistency across a full sequence.

\paragraph{DFM.} DFM ~\cite{tewari2023forwarddiffusion} incorporates a neural radiance field into the architecture of an image diffusion model such that novel views are 3D consistent by design. By employing pixelNeRF~\cite{pixelnerf}, DFM generates the 3D representation given a set of conditioning views. Starting from a single view, it generates an extrapolated target view that acts as additional conditioning in all subsequent sampling steps. 
\section{Runtime}
\label{sec:runtime}
In Tab.~\ref{tab:runtime}, we compare the runtimes of the evaluated methods for generating 80 frames of a video sequence on an NVIDIA RTX4090 GPU with 24GB VRAM. GenWarp achieves the fastest sampling time, as high-quality but inconsistent novel views can already be obtained with 20 DDIM steps.
Although MV-LDM generates multiple views at a time, which improves 3D consistency and uses 70 DDIM steps to achieve good image quality, it is only slightly slower than the single-view generation of GenWarp.
Both DFM and PhotoNVS are an order of magnitude slower due to slow volumetric NeRF rendering and many denoising steps, respectively. Our proposed metric MEt3R can be evaluated in only $95ms$ per image pair. 

\begin{table}[t]
\centering
\resizebox{\columnwidth}{!}{\begin{tabular}{l|cccc}
\toprule[1pt]
\multicolumn{1}{c}{} & \textbf{GenWarp} & \textbf{PhotoNVS} & \textbf{DFM} & \textbf{MV-LDM} \\
\midrule
Runtime (s)  &    \textbf{70}           &    7840           &        1020      & \underline{100}   \\

\bottomrule[1pt]
\end{tabular}}
\caption{\textbf{Runtime comparison}. We report the runtime in seconds for all the baselines for generating a full video sequence comprising 80 frames. MV-LDM and GenWarp~\cite{seo2024genwarp} achieve the fastest sampling, followed by DFM~\cite{tewari2023forwarddiffusion} and then PhotoNVS~\cite{Yu2023PhotoconsistentNVS}.}
\label{tab:runtime}
\end{table}

\begin{figure*}[t]
    \centering
    \includegraphics[width=1\linewidth]{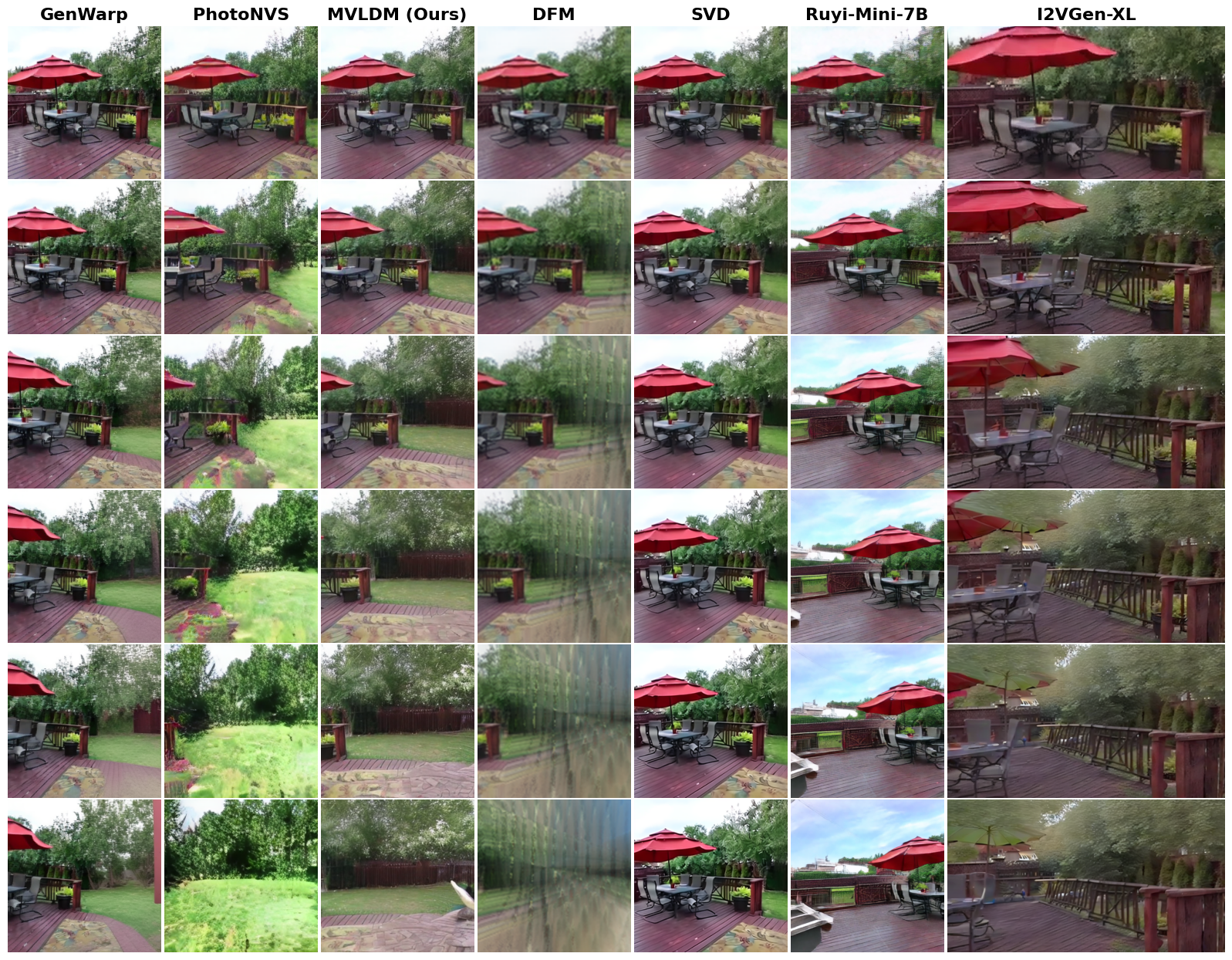}
    \caption{\textbf{Examples of generated multi-views and videos.} From Top $\rightarrow$ Down is the increasing frame number with columns for each method. Note that the first row is the input image, the first four columns are the results of multi-view generation models with explicit camera control, whereas the last three columns are generated videos from video diffusion models without any camera control.}
    \label{fig:qualitative_appendix_1}
\end{figure*}

\begin{figure*}[t]
    \centering
    \includegraphics[width=1\linewidth]{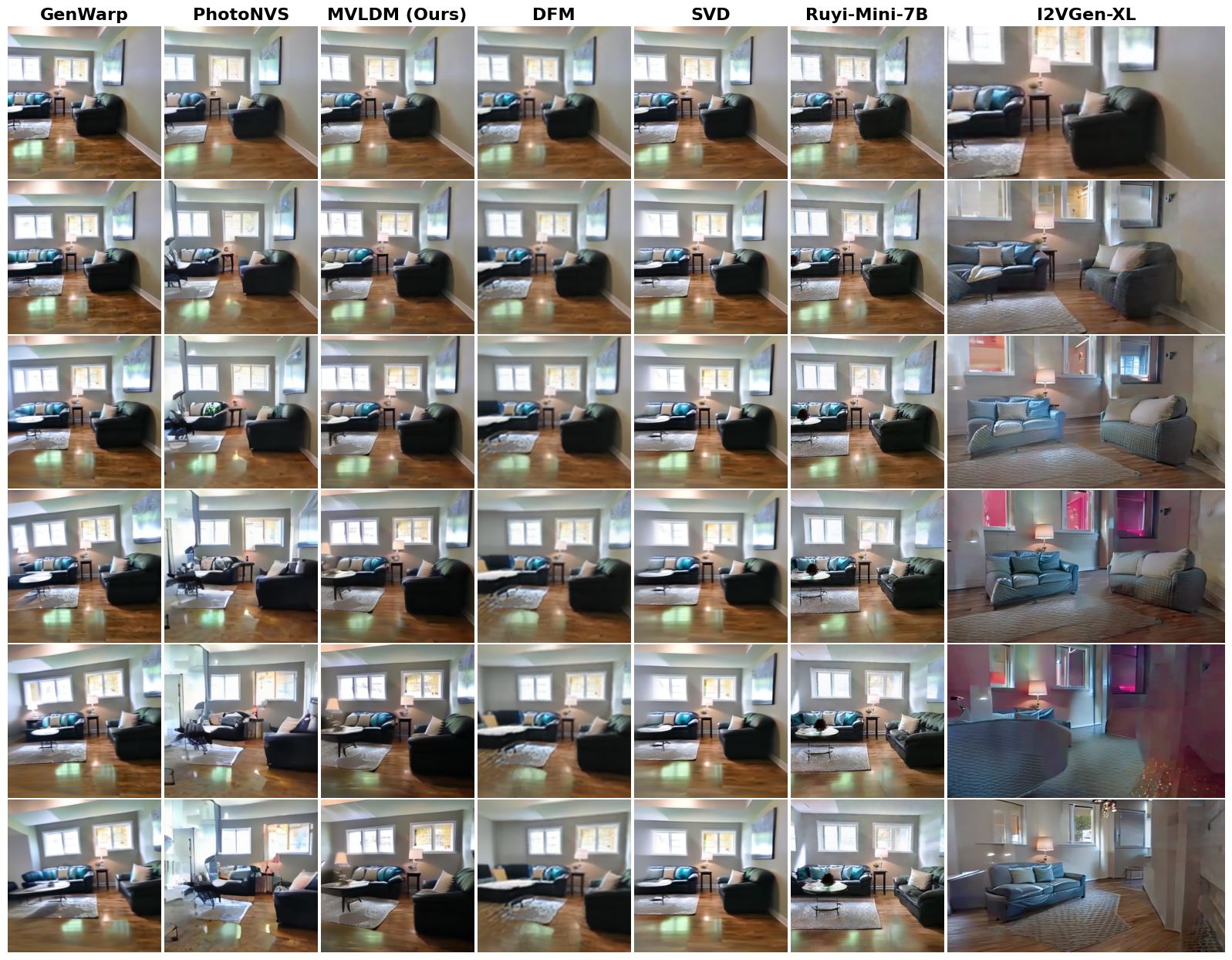}
    \caption{\textbf{Examples of generated multi-views and videos.} From Top $\rightarrow$ Down is the increasing frame number with columns for each method. Note that the first row is the input image, the first four columns are the results of multi-view generation models with explicit camera control, whereas the last three columns are generated videos from video diffusion models without any camera control.}
    \label{fig:qualitative_appendix_2}
\end{figure*}

\begin{figure*}[t]
    \centering
    \includegraphics[width=1\linewidth]{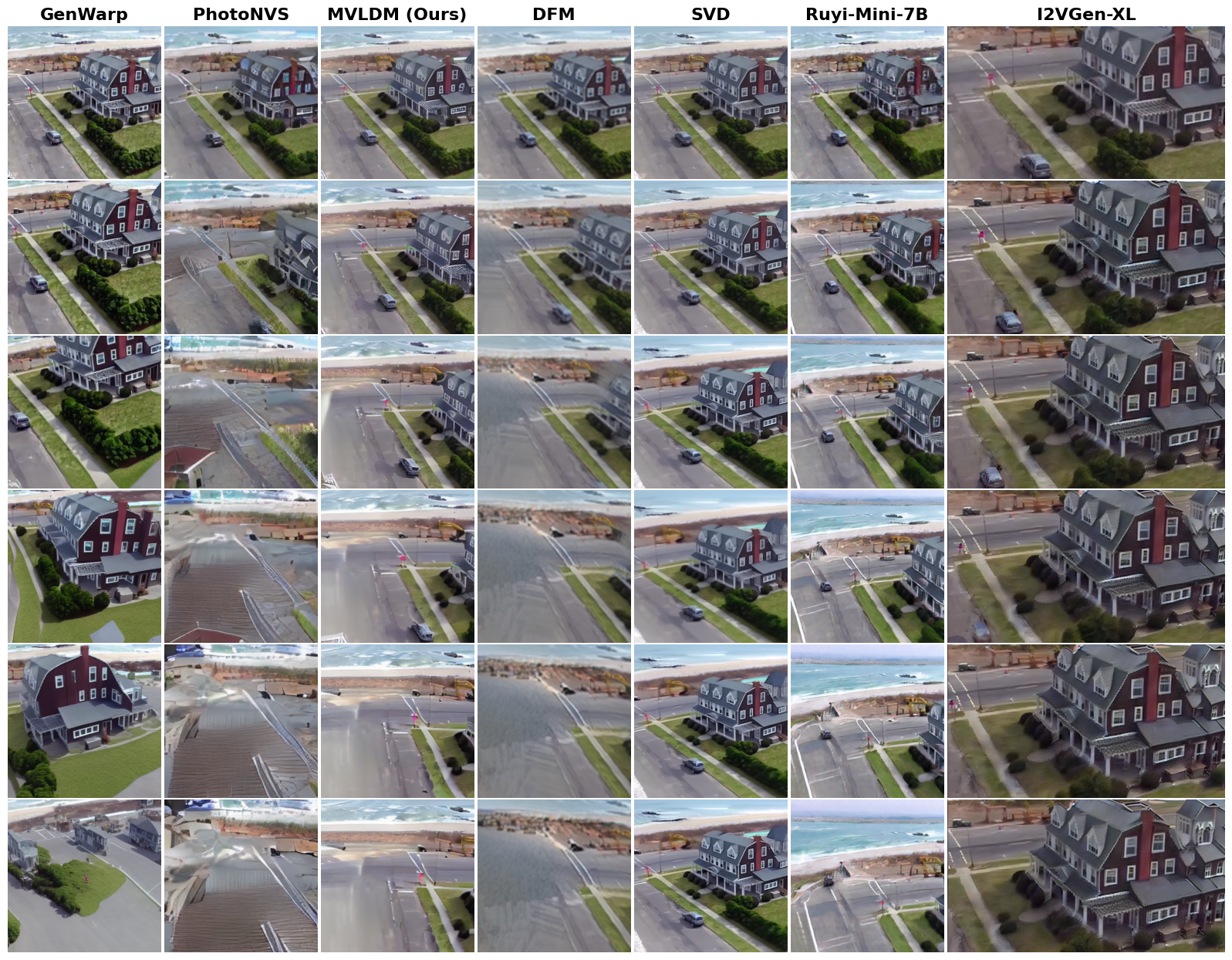}
    \caption{\textbf{Examples of generated multi-views and videos.} From Top $\rightarrow$ Down is the increasing frame number with columns for each method. Note that the first row is the input image, the first four columns are the results of multi-view generation models with explicit camera control, whereas the last three columns are generated videos from video diffusion models without any camera control.}
    \label{fig:qualitative_appendix_4}
\end{figure*}

\begin{figure*}[t]
    \centering
    \includegraphics[width=1\linewidth]{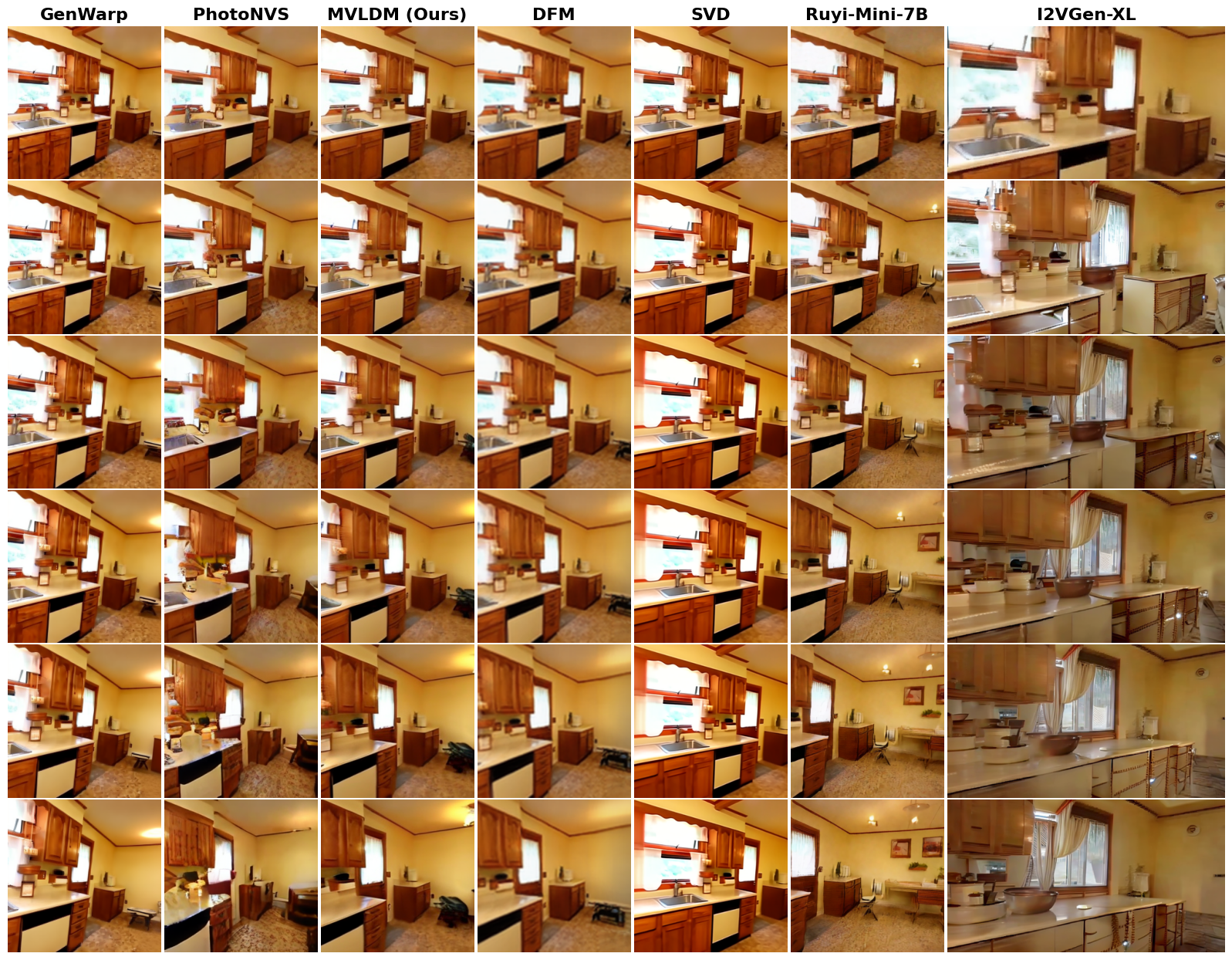}
    \caption{\textbf{Examples of generated multi-views and videos.} From Top $\rightarrow$ Down is the increasing frame number with columns for each method. Note that the first row is the input image, the first four columns are the results of multi-view generation models with explicit camera control, whereas the last three columns are generated videos from video diffusion models without any camera control.}
    \label{fig:qualitative_appendix_5}
\end{figure*}

\begin{figure*}[t]
    \centering
    \includegraphics[width=1\linewidth]{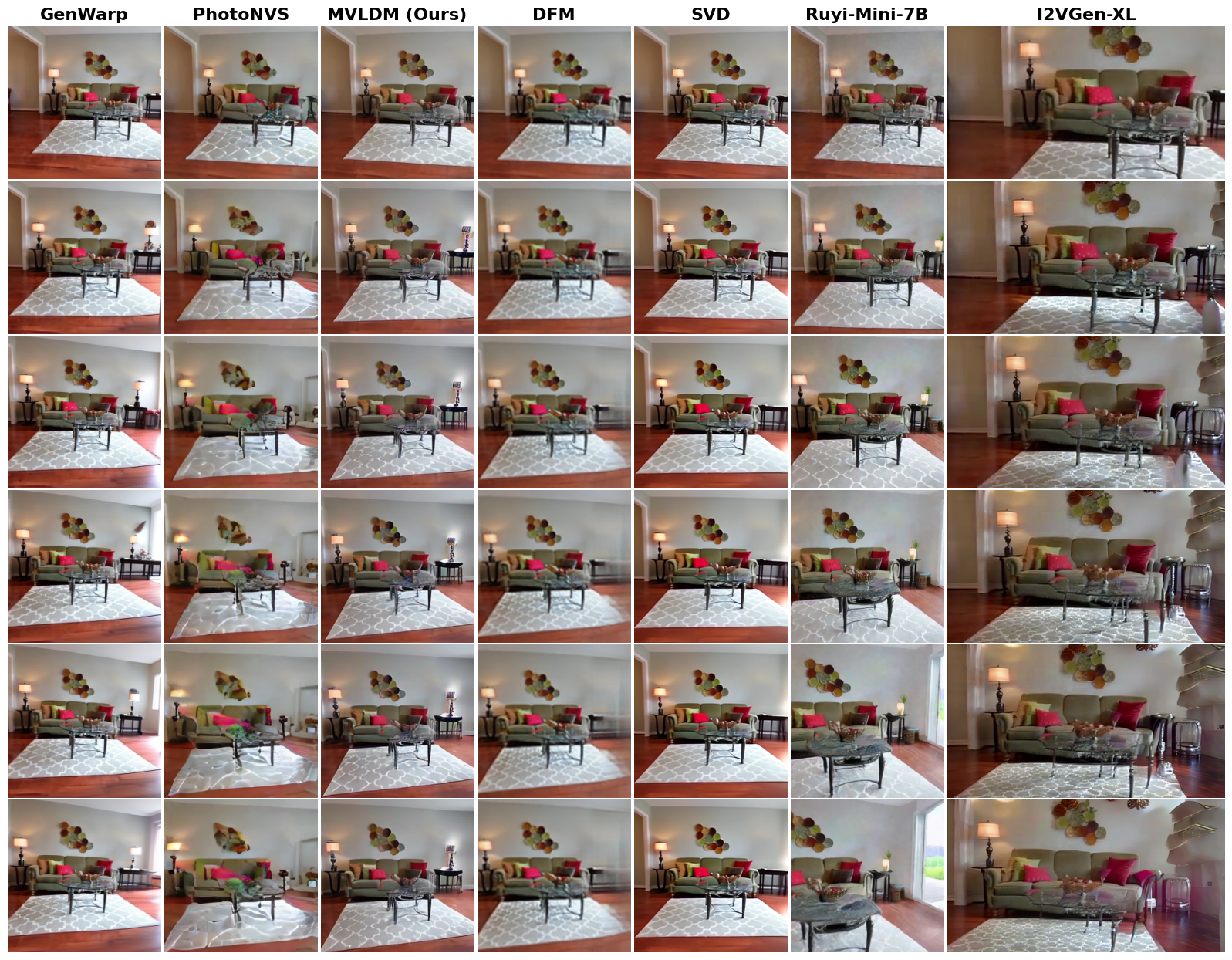}
    \caption{\textbf{Examples of generated multi-views and videos.} From Top $\rightarrow$ Down is the increasing frame number with columns for each method. Note that the first row is the input image, the first four columns are the results of multi-view generation models with explicit camera control, whereas the last three columns are generated videos from video diffusion models without any camera control.}
    \label{fig:qualitative_appendix_6}
\end{figure*}

\begin{figure*}[t]
    \centering
    \includegraphics[width=1\linewidth]{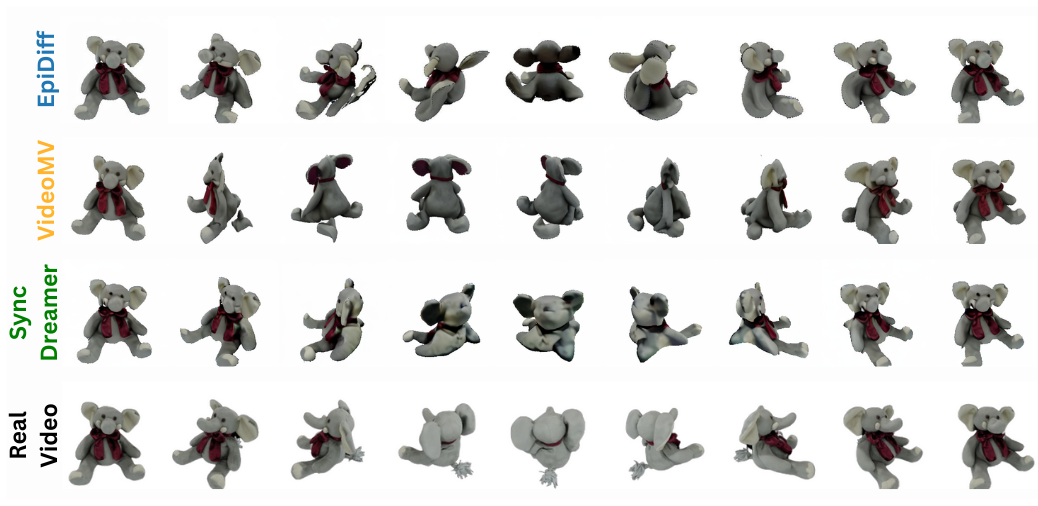}
    \caption{\textbf{Qualitative results on GSO.} A $360^\circ$ rendering of the GSO object \emph{Elephant}.}
    \label{fig:qualitative_obj_1}
\end{figure*}

\begin{figure*}[t]
    \centering
    \includegraphics[width=1\linewidth]{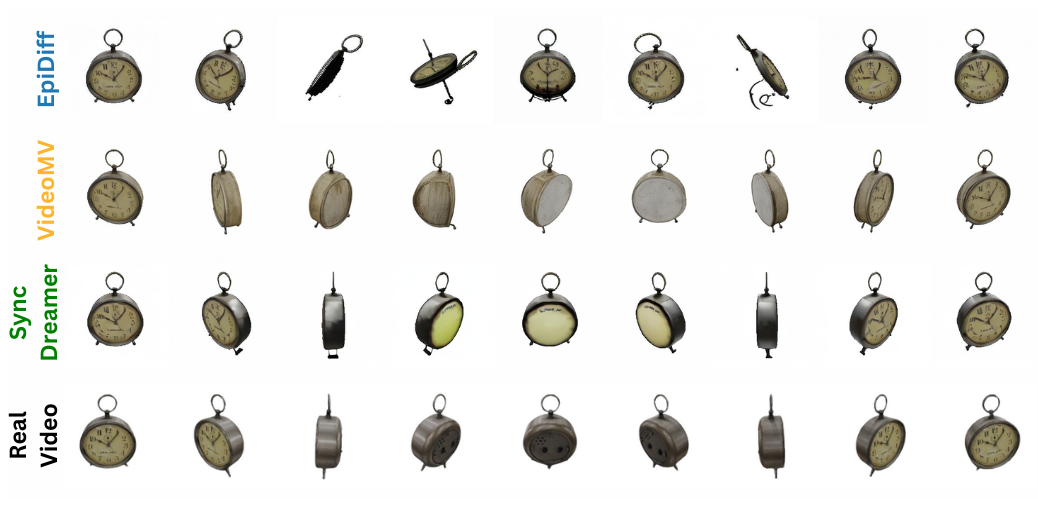}
    \caption{\textbf{Qualitative results on GSO.} A $360^\circ$ rendering of the GSO object \emph{Alarm}.}
    \label{fig:qualitative_obj_2}
\end{figure*}

\begin{figure*}[t]
    \centering
    \includegraphics[width=1\linewidth]{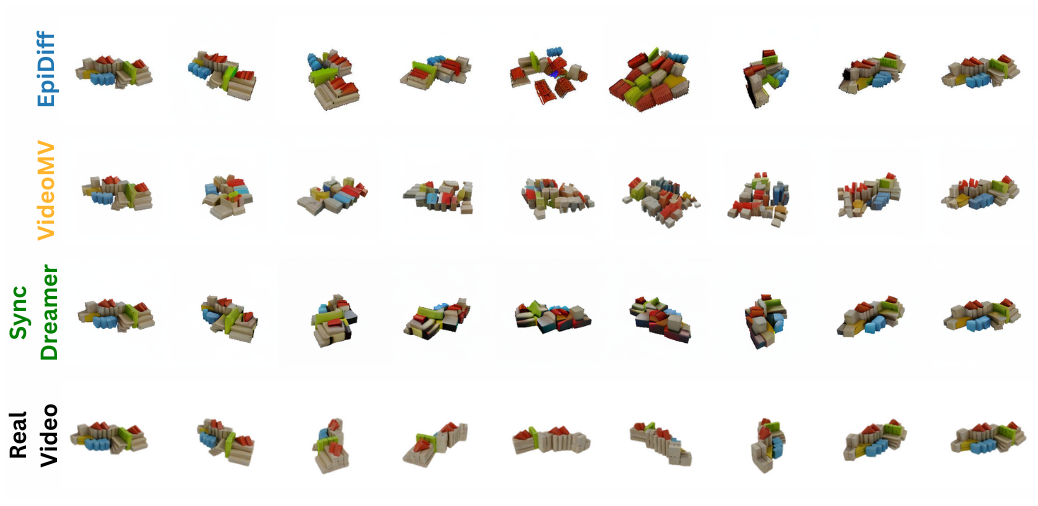}
    \caption{\textbf{Qualitative results on GSO.} A $360^\circ$ rendering of the GSO object \emph{Blocks}.}
    \label{fig:qualitative_obj_3}
\end{figure*}

\begin{figure*}[t]
    \centering
    \includegraphics[width=1\linewidth]{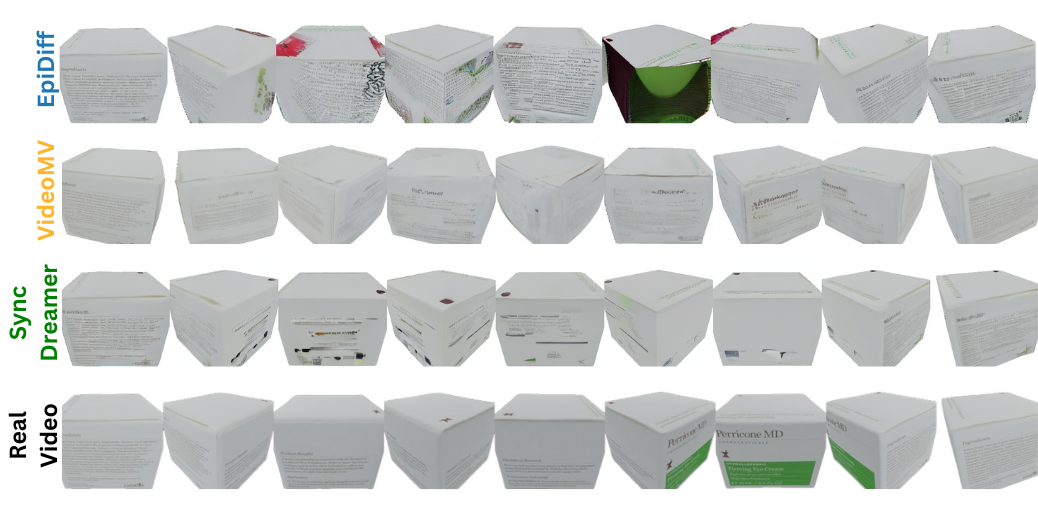}
    \caption{\textbf{Qualitative results on GSO.} A $360^\circ$ rendering of the GSO object \emph{Cream}.}
    \label{fig:qualitative_obj_4}
\end{figure*}

\end{appendices}